\documentclass[10pt,journal,compsoc]{IEEEtran}

\usepackage{paralist}
\usepackage{enumitem}
\usepackage{lineno}
\usepackage{tabularx}
\usepackage{graphicx}
\usepackage{hhline}
\usepackage{multirow,booktabs}
\usepackage{mathtools}
\usepackage{amssymb}
\usepackage{color}
\usepackage{subcaption}
\usepackage{url}
\usepackage{mathtools}
\usepackage[square,comma,numbers,sort&compress,sectionbib]{natbib}

\usepackage{color}

\usepackage[bookmarks=true%
,bookmarksnumbered=true%
,hypertexnames=false%
,breaklinks=true%
,colorlinks=true%
	,linkcolor=black
	,urlcolor=black
	,anchorcolor=black
	,citecolor=black
]{hyperref}

\usepackage{setspace}
\usepackage{tikz-qtree}

\usepackage{tikz}
\usetikzlibrary{calc,spy,shapes}
\usetikzlibrary{arrows,decorations.pathmorphing,fit,positioning,patterns}

\newcommand{\conditional}{\mspace{2mu} | \mspace{2mu}}
\newcolumntype{C}{>{\centering\arraybackslash}X}
\newcommand{\todo}[1]{ \textcolor{red}{\textbf{TODO: #1}}}

\DeclarePairedDelimiter\floor{\lfloor}{\rfloor}

\begin{document}

\title{HiTR: Hierarchical Topic Model Re-estimation\\ for Measuring Topical Diversity of Documents}

\author{Hosein~Azarbonyad,~
        Mostafa~Dehghani,~
        Tom~Kenter,~
        Maarten~Marx,~
        Jaap~Kamps,~
        and~Maarten~de~Rijke% <-this % stops a space
        \IEEEcompsocitemizethanks{ \IEEEcompsocthanksitem H. Azarbonyad, T. Kenter, M. Marx, and M. de Rijke are with Informatics Institute, University of Amsterdam.
	\IEEEcompsocthanksitem M. Dehghani and J. Kamps are with Institute for Logic, Language, and Computation, University of Amsterdam.\protect\\
E-mail: \{h.azarbonyad, dehghani, maartenmarx, kamps, derijke\}@uva.nl
\hspace*{1cm} tom.kenter@gmail.com        
        }% <-this % stops an unwanted space
}

\markboth{}%
{Azarbonyad \MakeLowercase{\textit{et al.}}: HiTR: Hierarchical Topic Model Re-estimation for Measuring Topical Diversity of Documents}

\newcommand{\rqone}{How effective is our hierarchical re-estimation approach in measuring topical diversity of documents? How does its effectiveness compare to the state-of-the-art in addressing the general and impure topics problem? Are the thus improved topic models also successfully applicable in other tasks?}
\newcommand{\rqtwo}{What is the effect of DR on the quality of topic models? Can DR replace manual pre-processing?}
\newcommand{\rqthree}{Does TR increase the purity of topics? And if so, how does using the more pure topics influence the performance in topical diversity task?}
\newcommand{\rqfour}{How does TAR affect the sparsity of document-topic assignments? And what is the effect of the achieved parsimonized document-topic assignments on the topical diversity task?}

\IEEEtitleabstractindextext{%
\begin{abstract}
%why is it interesting
A high degree of topical diversity is often considered to be an important characteristic of interesting text documents.
A recent proposal for measuring topical diversity identifies three distributions for assessing the diversity of documents: distributions of words within documents, words within topics, and topics within documents. 
%the problem
Topic models play a central role in this approach and, hence, their quality is crucial to the efficacy of measuring topical diversity.
The quality of topic models is affected by two causes: \emph{generality} and \emph{impurity} of topics. 
%
%General topics only include common information of a background corpus. Such topics are assigned to most of the documents.
General topics only include common information of a background corpus and are assigned to most of the documents.
Impure topics contain words that are not related to the topic. Impurity lowers the interpretability of topic models. Impure topics are likely to get assigned to documents erroneously. 
%The solution
We propose a hierarchical re-estimation process aimed at removing generality and impurity.
%Re-estimation is the process of extracting essential elements of a distribution and removing unnecessary information from it. 
Our approach has three re-estimation components: \begin{inparaenum}[(1)]
\item \emph{document re-estimation}, which removes general words from the documents;
\item \emph{topic re-estimation}, which re-estimates the distribution over words of each topic; and
\item \emph{topic assignment re-estimation},  which re-estimates for each document its distributions over topics.\end{inparaenum}\ 
%Say what your solution achieves
For measuring topical diversity of text documents, our HiTR approach improves over the state-of-the-art measured on PubMed dataset.
%Say what follows from your solution
%The effectiveness of HiTR is also evaluated on  two other tasks: document clustering and classification. 
%Moreover, the results show that HiTR is able to achieve a higher performance than other topic models in other tasks such as document clustering and classification.
\end{abstract}

}

\maketitle

\IEEEdisplaynontitleabstractindextext

% \fi
\IEEEpeerreviewmaketitle

% !TEX root = ./main.tex

%\todo{This is a fucking great introduction!!!!}

\IEEEraisesectionheading{\section{Introduction}\label{sec:introduction}}

\IEEEPARstart{Q}uantitative notions of measuring topical diversity of text documents are useful in a number of applications, such as assessing the interdisciplinariness of a research proposal~\cite{Bache2013} and helping to determine the interestingness of a document~\cite{Azarbonyad2015, Azarbonyad2017-ECIR, Derezinski2015}.

Well over three decades ago, an influential formalization of diversity was introduced in biology~\cite{Rao1982}. It decomposes diversity in terms of three central concepts: \emph{elements} that belong to \emph{categories} within a  \emph{population}~\cite{solow-measurement-1993}. Given a set $T$ of categories which partitions a population $d$, the diversity of   $d$ is then defined  as
\begin{equation}
\label{eq:topdiv}
div(d) = \sum_{i\in T} \sum_{j\in T} p_i^d p_j^d \delta(i,j),
\end{equation}
%\todo{Define $T$. And state whether the categories are disjoint? I guess they form a partition. If so, say that, otherwise explain!}
%\todo{I do not beieve it is expected distancem bevause you count double plus you take the "reflexive" points too. So this formula is not correct. Also explain what you mean by distnace between categories. Is the distance between 2 elements in 1 category equal to 0?}
where   $p_i^d$ denotes the  proportion of category $i$  in $d$ and  $\delta(i, j)$ is the distance between categories $i$ and $j$, which can be calculated in a chosen manner.   This notion of population diversity can be interpreted as the expected distance between two randomly selected (with replacement) elements of the population.

\citet{Bache2013} have adapted the biological notion of population diversity to quantify the topical diversity of a text document. For measuring the topical diversity of a text document, words are considered elements, topics are categories, and a document is a population. When using topic modeling for measuring topical diversity of a text document $d$, \citet{Bache2013} estimate elements based on the probability of a word given the document ($P(w\conditional  d)$), categories based on the probability of a word given a topic ($P(w\conditional  t)$), and populations based on the probability of a topic given the document ($P(t\conditional  d)$).

In probabilistic topic modeling, at estimation time, these distributions are usually assumed to be sparse. First, the main content of documents is assumed to be generated by a small subset of words from the vocabulary (i.e., $P(w\conditional  d)$ is sparse). Second, each topic is assumed to contain only some topic-specific related words (i.e., $P(w\conditional  t)$ is sparse). Finally, each document is assumed to deal with a few topics only (i.e., $(P(t\conditional  d)$ is sparse). When approximated using currently available methods, however, $P(w\conditional  t)$ and $P(t\conditional  d)$ often turn out to be dense rather than sparse \cite{Soleimani2015, Wallach2009, Lin2014}. Dense distributions cause two important problems for the quality of topic models: \emph{generality} and \emph{impurity}. General topics mostly contain general words. They are typically assigned to most of the documents in a corpus. In other words, the $P(t\conditional  d)$ distributions are not document-specific. Impure topics contain words that are not related to the topic. These impure words are mostly general words. Generality and impurity of topics are problematic when estimating topical diversity of text documents since they both result in low quality $P(t\conditional  d)$ distributions.  Recall that these are core to the topical diversity score based on the biological notion of diversity (Equation \ref{eq:topdiv}).

To improve the measurement of topical diversity of text documents we propose a hierarchical way of making the three distributions $P(w\conditional d)$, $P(w\conditional t)$ and $P(t\conditional d)$ more sparse.
To this end we re-estimate the parameters of these distributions so that general, collection-wide items are removed and only salient items are kept. For the re-estimation, we use the concept of parsimony~\cite{Hiemstra2004} to extract only essential parameters of each distribution.
%In this manner, we extract only essential parameters of each distribution and achieve high quality topic models to be applied to estimate topical diversity.

We evaluate the performance of the proposed hierarchical re-estimation method for measuring topical diversity of text documents and compare our approach against the state-of-the-art \citep{Soleimani2015}. In doing so, we answer our main \textbf{research question}:
\begin{itemize}
  \item[] How effective is our hierarchical re-estimation approach in measuring topical diversity of documents? How does its effectiveness compare to the state-of-the-art in addressing the general and impure topics problem? Are the thus improved topic models also successfully applicable in other tasks?
\end{itemize}
%\todo{Can't we add a third question stating something like "Are the thus improved topic models also succesfully applicable in other tasks?" Or, Should one use our resestimation metjod as a general preprocessing step when using topic models? Do you see what I mean? I want to indicate this wider applicability.}

\noindent%
Our main contributions are: \begin{inparaenum}[(1)]
\item We propose a hierarchical re-estimation process for topic models to address the two main problems in estimating the topical diversity of text documents, using a biologically inspired definition of diversity.
\item We study each level of re-estimation individually in terms of efficacy in solving the general topics problem, the impure topics problem, and improving the accuracy of estimating the topical diversity of documents.
\item We study the impact of re-estimation parameters on the statistics of documents and its relation to the quality of trained topic models and recommend effective settings of these parameters.
\end{inparaenum}

As an additional contribution, we also make the source code of our topic model re-estimation method available to the research community to further advance research in this area \footnote{The source code is available here: \url{https://github.com/HoseinAzarbonyad/HiTR}}.

% !TEX root = ./main.tex

%\todo{very very good overview of all relevant leiterature}

%\todo{Maybe you might enumerate a list of applications of topic models in which the generalsity and the impurity problem have a negative influence on the result of the application. In diversity this is rather clear, but for other tasks, it might not be immediate. This would correspond to Step 4 in Jaap's thing: what does your solution entail?}
\section{Related Work}
\label{RelatedWork}
Our hierarchical topic model re-estimation touches on research in multiple areas. We review work in four directions: improving the quality of topic models, measuring text diversity, evaluating topic models, and parsimonization.

\subsection{Improving the quality of topic models}
\label{relwork:ImprovingtheQualityofTopicModels}

Topic models are effective for modeling text documents and expressing the contents of text documents in a low-dimensional space \cite{Blei2003}. Although topic models like Latent Dirichlet Allocation (LDA) are powerful tools for modeling data in an unsupervised fashion, they suffer from different issues, especially when dealing with noisy data \cite{Boyd2014}. As mentioned already, the two most important issues with topic models are the \emph{generality problem} and the \emph{impurity problem} \cite{Wallach2009, Soleimani2015, Boyd2014, Lin2014}. These problems with topic models have a negative influence on the performance of tasks in which topic models are applied besides  document diversity, namely  document clustering, document classification, document summarization, information retrieval, sentiment analysis (see \citep{Boyd2014} for an overview).

\citet{Wallach2009} propose asymmetric Dirichlet priors to construct a general topic and assign general terms to this general topic in the learning process.
Similar ideas to improve the quality of topic models have been employed by others~\citep{Wang2009, Williamson2010}.
Similar to \cite{Wallach2009, Wang2009, Williamson2010}, one of our goals is to address the generality problem. 
The main difference, however, is that they do not aim to address the two issues mentioned with topic models directly and the topic representations and topic word distributions that they arrive at are neither parsimonious nor sparse. 
That is, in their approach, each topic could still have a non-zero assignment probability to each document. We hypothesize that parsimony is essential in topic modeling, since it is expected that each document only focuses on a few topics \cite{Soleimani2015} and in contrast to the work cited above our goal is to achieve this parsimony. 

\citet{Soleimani2015} propose parsimonious topic models (PTM) to address the generality and impurity problems.
A shared topic is created and general words are assigned to this topic. PTM achieves state-of-the-art results compared to existing topic models.
We also address the generality and impurity problems with topic models. 
%Unlike \cite{Soleimani2015}, however, instead of modifying the training procedure of LDA, we propose a method to refine the topic models.
The background language model in our model and the shared topic in PTM have similar functionalities. They both are used to handle and remove generality from topic-word distributions. However, in PTM, the shared topic is more complicated as for each word there are a few more parameters to be estimated: 
\begin{inparaenum}[(1)]
\item whether a word is topic-specific for each topic and 
\item probability of being topic-specific under each topic for each word.
\end{inparaenum}
In our approach, we model all this using a background language model with much fewer parameters. Moreover, we model and remove the generality in three different levels: document-word distribution, topic-word distribution, and document-topic distribution. 
PTM handles the generality in topic-word and document-topic distributions and does not handle the generality in document-word distribution explicitly.

\subsection{Evaluating topic models}
Topic models are usually evaluated either intrinsically, for example, in terms of their generalization capabilities, or extrinsically in terms of their contribution to external tasks \cite{Wallach2009b}. We focus on extrinsic evaluations of the effectiveness of our re-estimation approach. Our main evaluation concerns its effectiveness in measuring the topical diversity of text documents.
In addition, in Section~\ref{Analysis}, we analyze the effectiveness of our re-estimation approach in removing impurity from documents %in an extrinsic manner, viz.\ 
in terms of purity in document clustering and document classification tasks. 

Specifically, in the document classification task, topics are used as features of documents with values $P(t\conditional d)$. These features are used for training a classifier \cite{Nguyen2015, Lacoste2009, Soleimani2015}. In the document clustering task, each topic is considered a cluster and each document is assigned to its most probable topic \cite{Nguyen2015, Xie2013}. 
For the analyses in Section~\ref{Analysis}, following common practice (e.g., \cite{Nguyen2015, Mehrotra2013, Lau2014}), we use Purity and Normalized Mutual Information in the clustering task, and Accuracy as our prime evaluation metric in the classification task.
Furthermore, the quality of topic models can be measured by the quality of the term distributions per topic, in terms of topic coherence \cite{Lau2014, Nguyen2015}, and by having their interpretability judged by humans \cite{Chang2009, Newman2011}. 

\subsection{Text diversity and interestingness}
\label{relwork:TextDiversityandInterestingness}
Prior to~\citet{Bache2013}, measuring topical diversity of documents had not been studied comprehensively from a text mining perspective. 
\citet{Bache2013} use Rao's diversity score~(Equation \ref{eq:topdiv}) \cite{Rao1982} to quantify diversity of text documents by means of LDA topic models \cite{Blei2003}. In their framework, the diversity of a  document is proportional to the number of dissimilar topics it covers. Similar to \cite{Bache2013}, \citet{Derezinski2015} define the diversity of documents by means of topic models, but instead of Rao's measure they use an information theoretic diversity measure based on the Kullback Leibler divergence.
%centric measures and employ Jensen-Shanon divergence to measure the divergence of topics a document covers.
%\todo{Aha, here you come to my new hobby ;-) I also noticed the resemblance of Rao to a definition of entropy. Can you not go into this? I.e. if you define this distance in a information theoretic way in Rao, don't you get (aslmost) JS divergence?) E.g. if you would simply take log(Cat i/Cat j) as distance, this is not that crazy: as this means log(Cat i) - log(cat j), which is typically a simplictic way to calculate distance.} They also analyze the correlation of a text's topical diversity and its interestingness and found that these two indeed are highly correlated. %metrics have a high correlation.
%\todo{Homework: compute whether you can bring def 3 of these guys into the general Rao form of (1). If so, I would add it here. And maybe also mention this just below (1) then.}
\citet{Azarbonyad2015} also use Rao's diversity measure to quantify the diversity of political documents and analyze the correlation of topical diversity and interestingness over political documents. Their main finding, however, is different from \citet{Derezinski2015}'s conclusion, as they conclude that  although in general topical diversity and interestingness of political documents are somehow correlated, a text's topical diversity does not necessarily reflect its interestingness.

\subsection{Model parsimonization}
\label{ModelParsimonization}
Parsimonization refers to the process of extracting essential elements of a distribution and removing superfluous, general information. Parsimonization can be considered an unsupervised feature selection approach. The idea is to extract features containing information about samples and remove features that are not informative for explaining the samples \cite{Law2004, Constantinopoulos2006}.
Because our hierarchical re-estimation process builds on %insights about 
parsimonious language models (PLMs) \cite{Hiemstra2004}, we briefly review them. %PLMs.

PLMs were introduced in an information retrieval setting, in which %where 
language models are used to model documents as distributions over words. The goal of parsimonization in this context is to extract words that reflect the content of documents and remove collection-specific general words \cite{Dehghani2016-ICTIR}. 
To extract salient document-specific words for each document, some studies define a layered language model of documents where the language model of a document is composed of a general background model and a document-specific language model \cite{Zhai2001, Zhang2002, Dehghani2016-CLEF}.
The Expectation-Maximization (EM) algorithm is employed to estimate the parameters of such models.
Using this idea, \citet{Hiemstra2004} propose a method for parsimonizing document language models with the aim of removing general words by pushing the probabilities of the words that are well explained by the background model toward zero. We employ this approach for re-estimating and refining topic models.

Here we briefly recall the formal principles underlying PLMs. The main assumption is that the language model of a document is a mixture of its own specific language model and the language model of the collection:
\begin{equation}
\label{PLM-eq1}
P(w\conditional d) = \lambda P(w\conditional \tilde{\theta}_d) + (1-\lambda) P(w\conditional \theta_C),
\end{equation}
where $w$ is a term, $d$ a document, $\tilde{\theta}_d$ the document specific language model of $d$, $\theta_C$ the language model of the collection $C$, and $\lambda$ is a mixing parameter ($0\leq \lambda\leq 1$). The main goal is to estimate $P(w\conditional \tilde{\theta}_d)$ for each document.
Language model parsimonization is an iterative  EM algorithm in which the initial parameters of the language model are the parameters of the standard language model, estimated using maximum likelihood:
\begin{description}
\item[\normalfont\emph{Initialization}:] 
\[P(w\conditional \tilde{\theta}_d) = \frac{t{f}_{w, d}}{\sum_{w'\in d} t{f}_{w', d}},\]
\end{description}
where $tf_{w, d}$ is the frequency of $w$ in $d$. The following steps are done in each iteration of the algorithm:
\begin{description}
\item[\normalfont\emph{E-step}:] \mbox{}
\begin{equation}
\label{PLM-E-Step}
e_w = tf_{w, d} \cdot \frac{\lambda P(w\conditional \tilde{\theta}_{d})}{\lambda P(w\conditional \tilde{\theta}_{d}) + (1 - \lambda) P(w\conditional \theta_{C}))},
\end{equation}
\item[\normalfont\emph{M-step}:] \mbox{}
\begin{equation}
\label{eq3}
P(w\conditional \tilde{\theta}_{d}) = \frac{e_w}{\sum_{w'\in d} e_{w'}},
\end{equation}
\end{description}
where $\tilde{\theta}_{d}$ is the parsimonized language model of document $d$, which is initialized by the language model of $d$, $C$ is the background collection, $P(w\conditional \theta_{C})$ is estimated using maximum likelihood estimation, and $\lambda$ is a parameter that controls the level of parsimonization. A low value of $\lambda$ will result in a more parsimonized model while $\lambda=1$ yields a model without any parsimonization. The E-step gives high probability values to terms that occur relatively more frequently in the document than in the background collection, while terms that occur relatively more frequently in the background collection get low probability values. In the M-step the parameters are normalized to form a probability distribution again. After this step, terms that receive a probability lower than a predefined \textit{threshold} are removed from the model. The EM process will stop after a fixed number of iterations or when the models $\tilde{\theta}_{d}$ do not change significantly anymore.

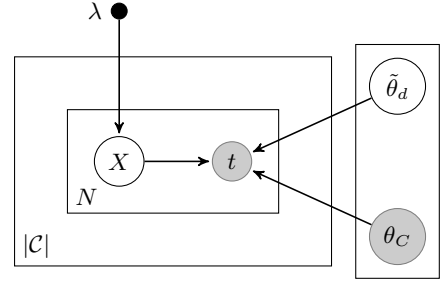
\begin{figure}[tp]
  \centering
  \begin{small}
  \begin{tikzpicture}
    [
      observed/.style={
      minimum size=1pt,circle,draw=black!50,fill=black!20
      },
      unobserved/.style={
      minimum size=1pt,circle,draw
      },
      hyper/.style={
      minimum size=1pt,circle,fill=black
      },
      post/.style={
      ->,>=stealth',shorten >=1pt,auto,node distance=7 cm,
                semithick, scale = 1, transform shape
      },
      %thick,scale=0.65, 
      %every node/.style={transform shape}
    ]

    \node (t) [observed] at (0,0) {$t$};
    \node (X) [unobserved] at (-1.5,0) {$X$};
    %\node (l) [unobserved] at (-3,0) {$\lambda$};
    \node (g) [unobserved] at (2.2,1) {$\tilde{\theta}_d$};
    \node (r) [observed] at (2.2,-1) {$\theta_{C}$};
    %\node (s) [observed] at (1.5,-1) {$\theta_s$};
    \node (q) [label=left:$\lambda$] at (-1.5,2) {};
    \filldraw [black] (-1.5,2) circle (3pt);
   
    \path
    (X) edge [post] (t)
    %(l) edge [post] (X)
    (r) edge [post] (t)
    %(s) edge [post] (t)
    (g) edge [post] (t)
    (q) edge [post] (X)
    ;

    \node [draw,fit= (t) (X), inner sep=30pt] (plate-context) {};
    \node [above right] at (plate-context.south west) {$|\mathcal{C}|$};
    \node [draw,fit=(t) (X), inner sep=10pt] (plate-token) {};
    \node [above right] at (plate-token.south west) {$N$};
    \node [draw,fit= (r) (g), inner sep=5pt] (plate-context) {};
    
  \end{tikzpicture}
  \end{small}
  \caption{\label{fig:graphical-model}Plate diagram of PLM. X corresponds to $e_w$ in Equation \ref{PLM-E-Step}.}
\vspace{-10pt}
\end{figure}

PLM is a two-topic mixture model (the graphical model is shown in Fig.~\ref{fig:graphical-model}, as can be seen $\theta_{C}$ is considered as an external observation and the goal is to estimate $\tilde{\theta}_{d}$ given $\theta_{C}$ and $\lambda$). In that sense, PLM is similar to an LDA model with two topics (general and specific topics). However, its mechanism is different than LDA. In LDA, all topics are shared among documents and only the proportions of topics (document-topic distributions) are different for different documents. In PLM, there is a general topic which is shared among all documents, but there is a specific topic for each document which is not shared with other documents. Moreover, in PLM, the $\lambda$ controls the proportion of general and specific topics in documents and it is fixed.

% !TEX root = ./main.tex

\section{Hierarchical Topic Model Re-estimation}
\label{MeasuringTopicalDiversityofDocuments}

%\todo{I like this to be more general than just for diversity} 
In this section, we describe HiTR (\emph{hi}erarchical \emph{to}pic model \emph{r}e-estimation).
HiTR can be applied on top of any topic modeling approach that has two main components, $P(w\conditional  t)$ and $P(t\conditional  d)$ distributions.
%, and that models documents as distributions over topics and topics as distributions over words. \todo{here you use twice the same thing: P(t|d) and mode;s as distributions oover topics. I don't get it and it reads hard}

\subsection{Overview}
\label{HiTR}
The input of HiTR is a corpus of text documents.  
The output is a probability distribution over topics for each document in the corpus. 
%We use this final distribution over topics to measure topical diversity of documents.

\begin{figure*}[t!]
\centering
\includegraphics[width=1.5\columnwidth]{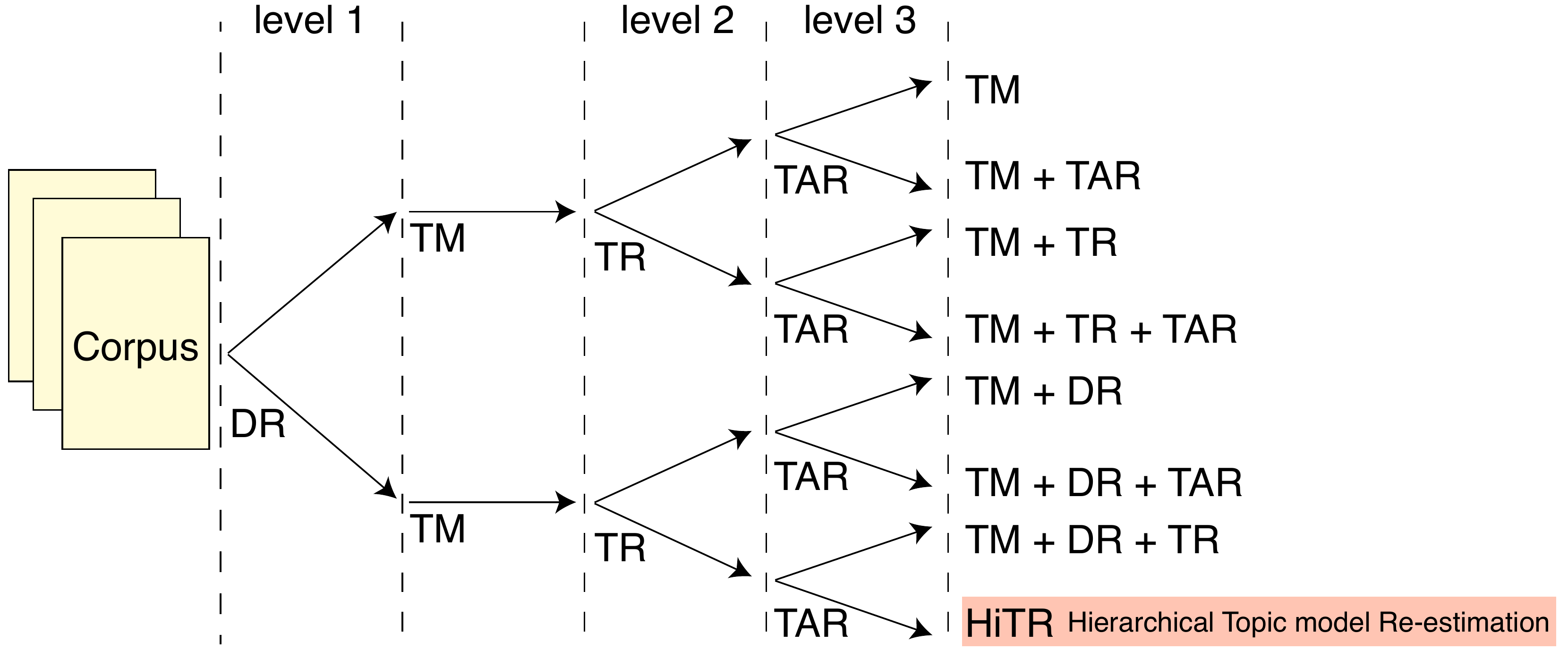}
\caption{Different topic re-estimation approaches. TM is a topic modeling approach like, e.g., LDA. DR is document re-estimation, TR is topic re-estimation, and TAR is topic assignment re-estimation. }
\label{figure:PTM}
\end{figure*}

As explained in the introduction, the quality of topic models such as LDA is highly dependent on the quality of the $P(w\conditional  d)$, $P(w\conditional  t)$, and $P(t\conditional  d)$ distributions. However, generality and impurity of these distributions cause the poor quality of topic models. % \todo{Again, try to get rid of this specific application here}
To solve these issues, we propose to apply re-estimation at three levels:
\begin{description}[nosep]
\item[document re-estimation (DR)] re-estimates the language model per document $P(w\conditional  d)$
\item[topic re-estimation (TR)] re-estimates the language model per topic $P(w\conditional  t)$
\item[topic assignment re-estimation (TAR)] re-estimates the distribution over topics per document $P(t\conditional  d)$\end{description}
Based on applying or not applying re-estimation at different levels, there are 8 possible re-estimation approaches.
Fig.~\ref{figure:PTM} gives a graphical overview of the different levels of re-estimation and how they are combined.
\emph{Hierarchical topic model re-estimation} (HiTR) refers to the model that uses all three re-estimation techniques, i.e., DR+TR+TAR that can be applied to any topic model TM. 

To summarize, HiTR works as follows: we first do the DR step, then train a topic model (TM step) on top of the re-estimated documents. Afterwards, we apply the TR step on the trained topic model and use the re-estimated topic model (the topic model achieved after TR step) to assign topics to documents. Finally, we apply the TAR step to topics assigned to the documents using the re-estimated topic model. We follow this order of re-estimation for two reasons: first, for the topical diversity task we only use the   document-topic distributions. And second,   this order provides the maximum amount of re-estimation in the final document-topic distribution because at each step of re-estimation  impurity and generality is removed from document-word and topic-word distributions and finally the remaining impurity and generality is removed using TAR. Next, we describe each of the re-estimation steps in more detail.

\subsection{Document re-estimation}
\label{DP}

The first level of re-estimation is \emph{document re-estimation} (DR), which re-estimates $P(w\conditional  d)$.
The main intuition behind this level of re-estimation is to remove unnecessary information from documents before training topic models. 
This is comparable to pre-processing steps such as removing stopwords and high and low frequency words, that are typically carried out prior to applying topic models \cite{Blei2003, Blei2012, Nguyen2015,Mehrotra2013,Lau2014}. 
Proper pre-processing of documents, however, takes lots of effort and involves tuning several parameters, such as the number of high-frequent words to remove, if stopwords should be removed or not, whether rare words should be removed or not, whether IDF values should be considered in removing general/rare words.
When dealing with a large document collection, finding optimum values for all of these parameters is non-trivial, while blindly removing words from documents without considering the distribution of them over documents could lead to missing important words and losing important information.

To solve this issue and pre-process documents automatically, we propose \emph{document re-estimation}. 
After document re-estimation, we can train any standard topic model on the re-estimated documents. If general words are absent from (re-estimated) documents, we expect that the trained topic models will not contain general topics. Moreover, 
document re-estimation removes impure elements (general words) from documents, which will lead to more pure topics.
Hence, document re-estimation is expected to %help 
address both the general topic and the impure topic problem.

Document re-estimation uses the parsimonization method described in \S\ref{ModelParsimonization}.
The parsimonized model $P(w\conditional  \tilde{\theta}_{d})$ in Equation~\ref{eq3} is used as the language model of document $d$, and after removing unnecessary words from $d$, the frequencies of the remaining words (words with $P(w\conditional  \tilde{\theta}_{d}) > 0$) are re-calculated for $d$ using the following equation:
\begin{equation*}
tf(w, d) = \floor*{P(w\conditional  \tilde{\theta}_{d}) \cdot |d|},
\end{equation*}
where $|d|$ is the document length in words. Topic modeling is then applied on the recalculated document-word frequency matrix.

\subsection{Topic re-estimation}
\label{TP}

The second level of re-estimation is \emph{topic re-estimation} (TR), which re-estimates $P(w\conditional  t)$ by removing general words from it.
The re-estimated distributions from this step are used to assign topics to documents.

The goal of this step is to increase the purity of topics by removing general words that have not yet been removed by document re-estimation.
It is known from the literature \cite{Wallach2009, Soleimani2015, Boyd2014,Lin2014} that some topics extracted by means of topic models are impure and contain general words. 
%Impure topics can lead to poor performance of topic models when used for measuring topical diversity of documents.

The two main advantages of applying TR are that \begin{inparaenum}[(1)] \item it results in more pure topics which are more interpretable by human, and \item after getting pure, topics are less likely to be assigned to documents erroneously.\end{inparaenum}  
%\todo{rewrite part 2, make "more document specific topic assignments a mathematical precise statement. This is too vague}

A topic is modeled as a distribution over words, which is itself a language model.
Our main assumption is that each topic's language model is a mixture of its topic-specific language model and the language model of the background collection.
\if 0
%
\begin{equation*}
P(w\conditional  t) = \lambda P(w\conditional  \tilde{\theta}_{t}) + (1-\lambda) P(w\conditional  \theta_T),
\end{equation*}
%
\fi
The goal of TR is to extract a topic-specific language model for each topic and remove the part which can be explained by the background model.
Given a set of topics $T$, background language model $\theta_T$, and for each $t\in T$, a topic-specific language model $\tilde{\theta}_{t}$, we initialize $P(w\conditional  \tilde{\theta}_{t})$ and $P(w\conditional  \theta_T)$ as follows:
%\todo{I am lost, you define completely different things than these two. Besides, below you tell what $\tilde{\theta}_{t}$ and $\theta_T$ are. So what is going on? This is not good.}
%
\begin{align*}
&P(w\conditional  \tilde{\theta}_{t}) = P(w\conditional  \theta_{t}^{\mathcal{TM}}) \\
&P(w\conditional  \theta_T) = \frac{\sum_{t \in T} P(w\conditional  \theta_{t}^{\mathcal{TM}})}{\sum_{w' \in V_T} \sum_{t' \in T} P(w'\conditional  \theta_{t'}^{\mathcal{TM}})}
\end{align*}

\noindent%
where   $P(w\conditional  \theta_{t}^{\mathcal{TM}})$ is the probability of $w$ belonging to topic $t$ estimated by a topic model $\mathcal{TM}$, and $V_T$ is the set of all words occurring in all topics.
Having these estimations, the steps of TR are similar to the steps of PLM, except that in the E-step we estimate $tf_{w,t}$ (the frequency of $w$ in $t$) using $P(w\conditional  \theta_{t}^{\mathcal{TM}})$.

\subsection{Topic assignment re-estimation}
\label{TAP}

The third and final level of re-estimation is \emph{topic assignment re-estimation (TAR)} which re-estimates $P(t\conditional  d)$.

In topic modeling, most topics are usually assigned with a non-zero probability to most of documents.
When documents are typically focused on just a few topics, this is an incorrect assignment, as topics should only be assigned to documents that deal with them. General topics assigned to a majority of documents are uninformative. 
The goal of TAR is to address the general topics problem and achieve more document specific topic assignments. 

To re-estimate topic assignments, a topic model is first trained on the document collection.
This model is used to assign topics to documents based on the proportion of words in common between them.
We then model the distribution over topics per document as a mixture of its document-specific topic distribution and the topic distribution of the entire collection.
%
\if 0
\begin{equation*}
P(t\conditional  d) = \lambda P(t\conditional  \tilde{\theta}_d) + (1-\lambda) P(t\conditional  \theta_C).
\end{equation*}
\fi
%
The goal of TAR is to extract the document-specific topic distribution for each document and remove general collection-wide topics from them.

We initialize the document-specific topic distribution $P(t\conditional  \tilde{\theta}_{d})$ and the distribution of topics in the entire collection $C$, $P(t\conditional  \theta_C)$ as follows:
\begin{align*}
 P(t\conditional  \tilde{\theta}_{d}) =~ & P(t\conditional  \theta_d^{\mathcal{TM}}) \\
 P(t\conditional  \theta_C) =~ & \frac{\sum_{d \in C} P(t\conditional  \theta_d^{\mathcal{TM}}) }{\sum_{t' \in T} \sum_{d' \in C} P(t'\conditional  \theta_{d'}^{\mathcal{TM}}) }.
\end{align*}
%
%\todo{Same as above, it is weird when you define something and afterwards you tell what you have defined, in between when you give meaning to other things. Of course it must be clear that $t$ is a topic and $d$ a document because in the first sentence of this paragraph, you have used already $P(t\mid d)$. So this is utterlu confusing, and bad math prctice}
%
Here  $P(t\conditional  \theta_d^{\mathcal{TM}}) $ is the probability of assigning topic $t$ to document $d$ estimated by the topic model $\mathcal{TM}$.
The remaining steps of TAR follow the ones of PLM. The only difference is that in the E-step, we estimate $f_{t,d}$ using $P(t\conditional  \theta_d^{\mathcal{TM}})$.

% !TEX root = ./main.tex

\section{Evaluating HiTR}\label{ExperimentalSetup}

To evaluate the performance of our approach to topical diversification, we follow the evaluation setup introduced in \cite{Bache2013}. Our main research question is:
\begin{description}
  \item[RQ1]
    \rqone
\end{description}
To address RQ1 we run our models on a binary classification task. We generate a synthetic dataset of documents with high and low topical diversity (the process is detailed in \S\ref{Dataset}), and the task for every model is to predict whether a document belongs to the high or low diversity class. We employ HiTR to re-estimate topic models and use the re-estimated models for measuring topical diversity of documents.
We compare our method to LDA (as also used  in \cite{Bache2013} for the same purpose) and to the state-of-the-art parsimonious topic models PTM~\cite{Soleimani2015}. 
The results of experiments regarding RQ1 are discussed in \S\ref{TopicalDiversityResults}. Moreover, we evaluate the performance of HiTR in document clustering and classification tasks and analyze its effectiveness in these tasks. The results of these experiments are described in \S\ref{Analysis}.

Additionally, to gain deeper insights into how HiTR performs, we conduct a separate analysis of each level of re-estimation, DR, TR and TAR and answer the following research questions:

\begin{description}
  \item[RQ2] 
  \rqtwo
  \item[RQ3] 
  \rqthree
  \item[RQ4]
  \rqfour
\end{description}

RQ2 concerns the effectivenes of DR in removing general words from documents and its effect on the quality of topic models. To answer RQ2, we train LDA models with and without manual pre-processing and with and without DR. We compare the quality of models achieved using different combinations. This will show how effective is DR in pre-processing documents automatically.  
Moreover, we measure corpus statistics such as vocabulary size, average type-token ratio, average document length after running DR with different parameters. We train LDA models on the corpora achieved with different parameters and measure the quality of  trained models. Then, we analyze the correlation of corpus statistics achieved from DR with different parameters and the quality of models trained on them.
In \S\ref{DRResults}, the results regarding RQ2 are described. 

To answer RQ3, we first evaluate the performance of TR on the topical diversity task and compare its performance to DR and TAR.
We focus on its effectiveness in removing impure words from topics and perform a qualitative analysis on topic models before and after running TR.
The results of experiments regarding RQ3 are discussed in \S\ref{TRResults}.

To answer RQ4, we first evaluate TAR together with LDA in a topical diversity task and analyze its effect on the performance of LDA to study how successful TAR is in removing general topics from documents. The results of this experiment are presented in \S\ref{TARResults}.

\section{Topical Diversity with HiTR}
In this section, we discuss the experimental setup for the topical diversity test. 
\subsection{Topical Diversity Measure}
\label{TopicalDiversityMeasure}

After re-estimating words distributions in documents, topics, and document-topic distributions using HiTR, we use the final distributions over topics per document for measuring topical diversity. 
Diversity of texts is computed using Rao's coefficient (Equation \ref{eq:topdiv}).
\if 0
We recall that, for a document $d$, Rao's diversity coefficient is defined as:
\begin{equation*}
div(d) = \sum_{i=1}^{T} \sum_{j=1}^{T} p_i^d p_j^d \delta(i,j),
\end{equation*}
where $T$ is the set of extracted topics, $p_i^d$ is the probability of assigning topic $i$ to document $d$, and  $\delta(i,j)$ is the distance between topics $i$ and $j$, based on their co-occurrence over the documents.
\fi
For each topic $x$, observed in corpus $C$, we construct a vector $V_x$ of length $|C|$ (the number of documents in the corpus).
Each entry of this vector corresponds to a document $d_y$ and its value is assigned as: $V_x[y] = p_x^y$.
We use the normalized angular distance for measuring the distance between topics, since it is a proper distance function \cite{Azarbonyad2015}:
%To compute the normalized angular distance $\delta$ between topics $i$ and $j$, we put:
\begin{equation*}
\delta(i,j) = \frac{ArcCos(CosineSim(V_i,V_j))}{\pi},
\end{equation*}
where $CosineSim(\cdot,\cdot)$ is the cosine similarity between two vectors, and $ArcCos(\cdot)$ is the arc cosine. 
We use the distributions over topics per document for calculating the distance between topics. 
There are two possible approaches for measuring the topic distance: based on document-topic distributions or topic-word distributions. From a diversity perspective, document-topic distributions are more suitable for this task. For example, consider two topics which co-occur frequently in documents but have different topic-word distributions.  In principle, if a document contains these topics, it should not be diverse, but since the  topic-word similarity of these two topics is low the document will have a high diversity.

\subsection{Dataset}
\label{Dataset}

We use the PubMed abstracts dataset \cite{PubMed} in our experiments.
This dataset contains articles published in bio-medical journals. We use the articles published between 2012 to 2015 for training topic models. This subset contains about 300,000 documents.
Following~\cite{Bache2013}, we generate 500 documents with a high value of diversity and 500 documents with a low value of diversity. 
We create high diversity documents as follows: we first randomly select 10 pairs of journals. Each pair contains two journals that are relatively unrelated to each other (we select 20 journals in total).
For each pair of journals $A$ and $B$, we select 50 articles to create 50 new probability distributions over topics as follows:
we randomly select one article from $A$ and one article from $B$ and generate a document by averaging the selected articles' bag of topic counts. In this way, for each pair of journals we generate 50 documents with a high diversity value.
We create low diversity documents as follows: for each of the chosen 20 journals,  we perform a similar procedure but instead of choosing articles from two different journals, we select them from the same journal and generate 25 non-diverse documents.
In the final set we have 500 diverse and 500 non-diverse documents. 

\subsection{Baselines}
Our baseline for the topical diversity task is the method proposed in \cite{Bache2013}, which uses LDA for measuring topical diversity of documents.
As an additional baseline, we use PTM~\cite{Soleimani2015} instead of LDA for measuring topical diversity. PTM is the state-of-the-art in topic modeling approaches, and based on our results PTM is more effective than the method proposed in \cite{Bache2013}. Thus, PTM is our main baseline in this task.

\subsection{Metrics}

To measure the performance of topic models in the topical diversity task, we follow \cite{Bache2013} and report ROC curves and AUC values.
As another evaluation measure, we report the \emph{sparsity} of topic models: the average number of topics assigned to the documents of a corpus \cite{Soleimani2015}. This measure reflects the ability of topic models to achieving sparse $P(t\conditional d)$ distributions.
We also measure the \emph{coherence} of the extracted topics. This measure indicates the purity of $P(w\conditional t)$ distributions and a high value of coherence implies high purity within topics. For estimating the coherence of a topic model we use a reference corpus. As our reference corpus, we use a version of English Wikipedia.\footnote{We use a dump of June 2, 2015, containing 15.6 million articles.} We estimate the coherence of a topic model using normalized pointwise mutual information between the top $N$ words within a topic using the following equation \cite{Lau2014, Nguyen2015}:
\begin{equation}
\label{pmi-eq}
\mathit{NPMI}(T) = \sum_{t \in T} \sum_{w_i, w_j \in topN(t) \wedge i< j} \frac{\log \frac{P(w_i, w_j)}{P(w_i)P(w_j)}}{-\log(P(w_i, w_j))},
\end{equation}
where $T$ is the set of extracted topics, $topN(t)$ is the top $N$ most probable words within topic $t$. $w_i$ is a word, $P(w_i, w_j)$ is estimated based on the number of documents in which  $w_i$ and $w_j$ co-occur   divided by the number of documents in the reference corpus. $P(w_i)$ is estimated similarly, using maximum likelihood estimation. 

\subsection{Preprocessing}
\label{Preprocessing}
We first lowercase all the text in the corpus. Then, we remove the stopwords included the standard stop word list from Python's NLTK package.
In addition, we remove the 100 most frequent words in the collection and words with fewer than five occurrences.

\subsection{Model parameters}

As noted above, the topic modeling approach used in our experiments with HiTR is LDA.
Following~\cite{Bache2013, Soleimani2015, Roder2015} we set the number of topics to 100.
We set the two hyperparameters to $\alpha=1/T$ and $\beta=0.01$, where $T$ is the number of topics, following~\cite{Nguyen2015}.
In the re-estimation process, at each step of the EM algorithm, we set the threshold for removing unnecessary components from the model to $0.01$ and remove terms with an estimated probability less than this threshold from the language models, as in \cite{Hiemstra2004}.

We perform 10-fold cross validation, using 8 folds as training data, 1 fold as development set to tune the parameters, and 1 fold for testing.

\subsection{Statistical significance}

For statistical significance testing, we compare our methods to PTM using   paired two-tailed t-tests with Bonferroni correction. 
To account for multiple testing, we consider an improvement significant  if: $p \leq \alpha/m$, where $m$ is the number of conducted comparisons and $\alpha$ is the desired significance. We set $\alpha=0.05$. 
In \S\ref{results}, $^\blacktriangle$ and $^\blacktriangledown$ indicate that the corresponding method performs significantly better and worse than PTM, respectively.

% !TEX root = ./main.tex

\section{Results}
\label{results}
In this section, we first present the results of HiTR in topical diversity task. Then, we analyze each individual level of re-estimation.

\subsection{Topical diversity results}
\label{TopicalDiversityResults}

\begin{figure}[h]
\centering
\includegraphics[width=\columnwidth]{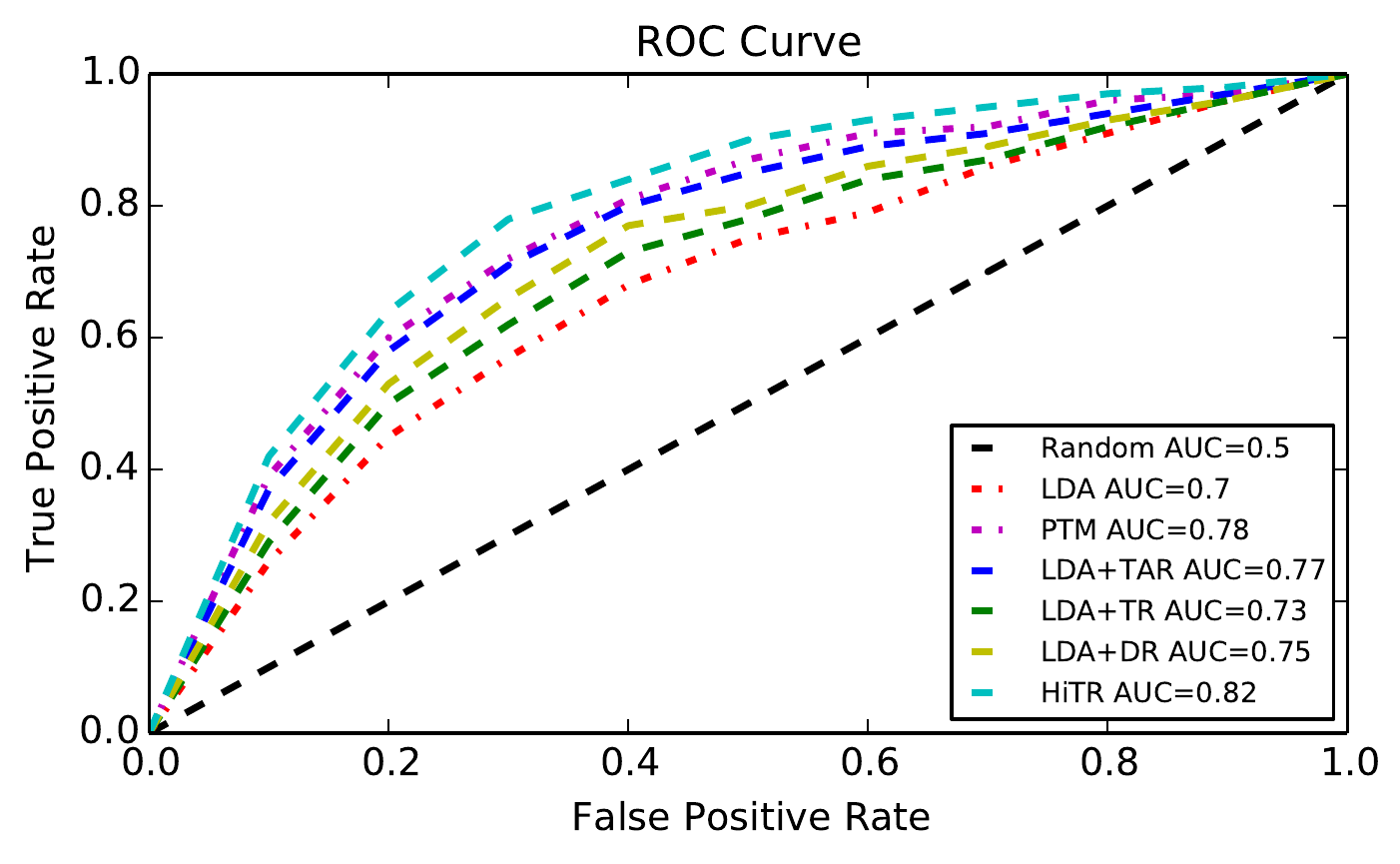}
\caption{Performance of topic models in topical diversity task on the PubMed dataset. The improvement of HiTR over PTM is statistically significant ({p-value}$<$0.05) in terms of AUC.}
\label{figure1}
\end{figure}

Fig.~\ref{figure1} plots the performance  of our topic models across different levels of re-estimation, and the models we compare to, on the Pub\-Med dataset. We plot ROC curves and compute AUC values. To plot the ROC curves we use the diversity scores calculated for the generated pseudo-documents with diversity labels.
HiTR improves the performance of LDA by 17\% and PTM by 5\% in terms of AUC. From Fig.~\ref{figure1} two observations can be made.

First, HiTR benefits from the three re-estimation approaches it encapsulates by successfully improving the quality of estimated diversity scores.
Second, the performance of LDA+TAR, which tries to address the generality problem, is higher than the performance of LDA+TR, which addresses impurity. General topics have a stronger negative effect on measuring topical diversity than impure topics. Also, LDA+DR outperforms LDA+TR. So, removing impurity from $P(t\conditional d)$ distributions is the most effective approach in the topical diversity task, and removing impurity from $P(w\conditional d)$ distributions is more effective than removing impurity from $P(w\conditional t)$ distributions.
Table \ref{table:divexample} illustrates the difference between LDA and HiTR with the topics assigned by the two methods for a non-diverse document that is combined from two documents from the same journal, entitled ``Molecular Neuroscience: Challenges Ahead'' and ``Reward Networks in the Brain as Captured by Connectivity Measures'', using the procedure described in \S\ref{Dataset}. As only a very basic stopword list was applied, words like \textit{also} and \textit{one} still appear. We expect to have a low diversity value for 
\begin{table}[t]
	\centering 
	\caption{\label{table:divexample}
          Topic assignments for a non-diverse document using LDA and HiTR. Only topics with $P(t\conditional d) > 0.05$ are shown.} 
	\begin{tabular}{@{~}c@{~~}c@{~~}l@{~}}
    \toprule
    \multicolumn{3}{c}{LDA}\\
    \midrule
    Topic & $P(t\conditional  d)$ & Top 5 words \\
    \midrule
    1 & 0.21 & brain, anterior, neurons, cortex, neuronal \\
    2 & 0.14 & channel, neuron, membrane, receptor, current \\
    3 & 0.10 & use, information, also, new, one \\
    4 & 0.08 & network, nodes, cluster, functional, node \\
    5 & 0.08 & using, method, used, image, algorithm \\
    6 & 0.08 & time, study, days, period, baseline \\
    7 & 0.07 & data, values, number, average, used \\
    \\
    \multicolumn{3}{c}{HiTR}\\
    \midrule
    Topic & $P(t\conditional  d)$ & Top 5 words \\
    \midrule
    1 & 0.68 & brain, neuronal, neurons, neurological,  nerve\\
    2 & 0.23 & channel, synaptic, neuron, receptor, membrane \\
    3 & 0.09 & network, nodes, cluster, community, interaction \\
    \bottomrule
    \end{tabular}
\end{table}
%
%
%As the documents are similar in content, 
the combined document. However, using Rao's  diversity measure, the topical 
diversity of this document based on the LDA topics is 0.97.
This is due to the fact that there are three document-specific topics---topics 1, 2 and 4---and four general topics. Topics~1 
and~2 are very similar and the $\delta$ between them is 0.13. The $\delta$ between, the other, more general topics is high; the average $\delta$ value between pairs of topics is as high as 0.38.
For the same document, HiTR only assigns three document-specific topics and they are more pure and coherent. 
%For example LDA topic 2 corresponds to topic to HiTR topic 2 in Table \ref{table:divexample}. General words such as ``current'' in the LDA topic are replaced by more salient words like ``synaptic'' in the HiTR topic. 
The average $\delta$ value between pairs of topics assigned by HiTR is 0.19. The diversity value of this document using HiTR is 0.16, which indicates that this document is non-diverse. 
%
%Hence, HiTR is more effective than other approaches in measuring topical diversity of documents; it successfully removes generality from $P(t\conditional  d)$.

Next, Table \ref{table:1.2} shows the sparsity of $P(t\conditional  d)$ using different topic models. 
All topic models that have TAR level of re-estimation achieve very sparse topic models. Thus, TAR contributes more to the sparsity achieved by HiTR. TAR increases the sparsity of LDA by more than 80\%.  
This sparsity leads to improvements over the performance of LDA on the topical diversity task, which indicates that TAR is able to remove general topics from documents. 
Topic models achieved by PTM are slightly more sparse than those achieved by HiTR. However, the difference is not statistically significant. The fact that HiTR outperforms PTM indicates that PTM extremely parsimonizes documents and throws away essential information from documents while HiTR removes mostly non-essential information from documents. 
%TAR also outperforms TR, indicating that \emph{general topics} have a stronger negative effect than \emph{impure topics} on the quality of topic models and estimated topical diversity scores. 
%
\begin{table}[h]
	\centering
	\caption{\label{table:1.2}
          Sparsity of topic models trained on PubMed for the topical diversity task. For significance tests we consider {p-value} $<$ 0.05/7.
          } 
	\begin{tabularx}{0.7\columnwidth}{l C}
          \toprule
          \textbf{Method}& \textbf{Sparsity}\\
          \midrule
          LDA & 13.77\hphantom{$^\blacktriangledown$} \\
          PTM& \textbf{\hphantom{0}1.78}\hphantom{$^\blacktriangledown$}\\
          \midrule
          LDA+DR & 13.17$^\blacktriangledown$\\
          LDA+TR &12.35$^\blacktriangledown$ \\
          LDA+TAR & 2.12\\
          LDA+DR+TR & 11.46$^\blacktriangledown$ \\
          LDA+DR+TAR & 2.01\\
          LDA+TR+TAR & 1.92\\
          \midrule
          HiTR & 1.80\\
          \bottomrule
	\end{tabularx}
\end{table}

\subsection{HiTR results}
In this section we analyze different levels of re-estimation to get insights on how different levels on re-estimation work individually and how much they are successful in removing non-necessary information from documents, topics, and topic-assignments.
\subsubsection{Document re-estimation results}
\label{DRResults}
\if 0
\textcolor{green}{
\begin{itemize}
\item We want to evaluate how DR affects the quality of topic models. 
\item We want to analyze the impact of DR parameters on the size of corpus and how it is correlated with the performance of topic models. 
\item We want to answer the following RQ: What is the effect of DR on the quality of topic models? Can DR replace the manual pre-processings done for improving the quality of topic models?
\item \todo{Discussion: We still keep a lot of rare words.}
\item \todo{Add Maarten's suggested experiment.}
\end{itemize}
}
\textcolor{red}{
\begin{itemize}
\item We train LDA models with and without manual pre-processing and with and without DR. We compare the quality of models achieved using different combinations. This will show how effective is DR in pre-processing documents automatically.  
\item We measure corpus statistics such as vocabulary size, average type-token ratio, average document length after running DR with different parameters. We train LDA models on the corpora achieved with different parameters and measure the quality of the trained models in terms of mutual information. Then we analyze the correlation of corpus statistics achieved from DR with different parameters and the quality of models trained on them.
\end{itemize}
}
\fi

In this section we focus on answering our second research question: What is the effect of DR on the quality of topic models? Can DR replace manual pre-processings? 

DR outperforms LDA by 7\% in measuring documents' topical diversity in terms of AUC. It also outperforms TR in this task but the difference is not significant. In fact, DR and TR are addressing the same problem with topic models. Both are successful in addressing \emph{impure topics}. However they are not successful in addressing the \emph{general topics} problem, since they have high value of sparsity.  

To analyze the effectiveness of DR in re-estimating documents and addressing the problems with topic models, we design an experiment in which no manual pre-processing is done and topic models are trained on these not-pre-processed documents. Our expectation is that even without doing any pre-processing a method that addresses the generality problems with topic models should still be able to achieve a good performance and do the pre-processing implicitly and automatically. Since DR tries to pre-process documents automatically, it should achieve a high quality topic model on these datasets. Table \ref{tab:DRCoherence} shows the performance of LDA, DR, and LDA+DR+TR in terms of their coherence. As expected, the coherence of LDA decreases by more than 23\% when no pre-processing is done on documents. More interestingly, adding DR scores better, both in terms of coherence and AUC, than manual pre-processing.
%This shows that manual pre-processing has a great impact on the quality of LDA and without doing a proper manual pre-processing the quality of extracted topic models are very low.  
%DR achieves a higher coherence compared to LDA even without pre-processing. This shows that DR is able to automate the manual pre-processing needed before training topic models. 

\begin{table}[t]
	\centering % centering table
	\caption{\label{tab:DRCoherence}
	The effect of document pre-processing on the quality of topic models measured in terms of coherence and AUC achieved in the 			topical diversity task. 
	%Based on conducted one-tailed t-test significance test, all the differences are statistically significant ($p-value<0.05$)
	} 
	\begin{tabular}{l l l}
    %\toprule
    \hline
   \textbf{Method}& $Coherence$  & $AUC$ \\
   \hline
   LDA (without pre-processing) &  6.23 & 0.54 \\
   LDA+pre-processing &   8.45 & 0.73\\
   LDA+DR &  8.95 & 0.75\\
   LDA+DR+TR &  10.29 & 0.79\\
	\hline
	\end{tabular}
\end{table}

\begin{figure*}[ht] 
  \begin{subfigure}[b]{0.5\linewidth}
    \centering
    \includegraphics[width=0.75\linewidth]{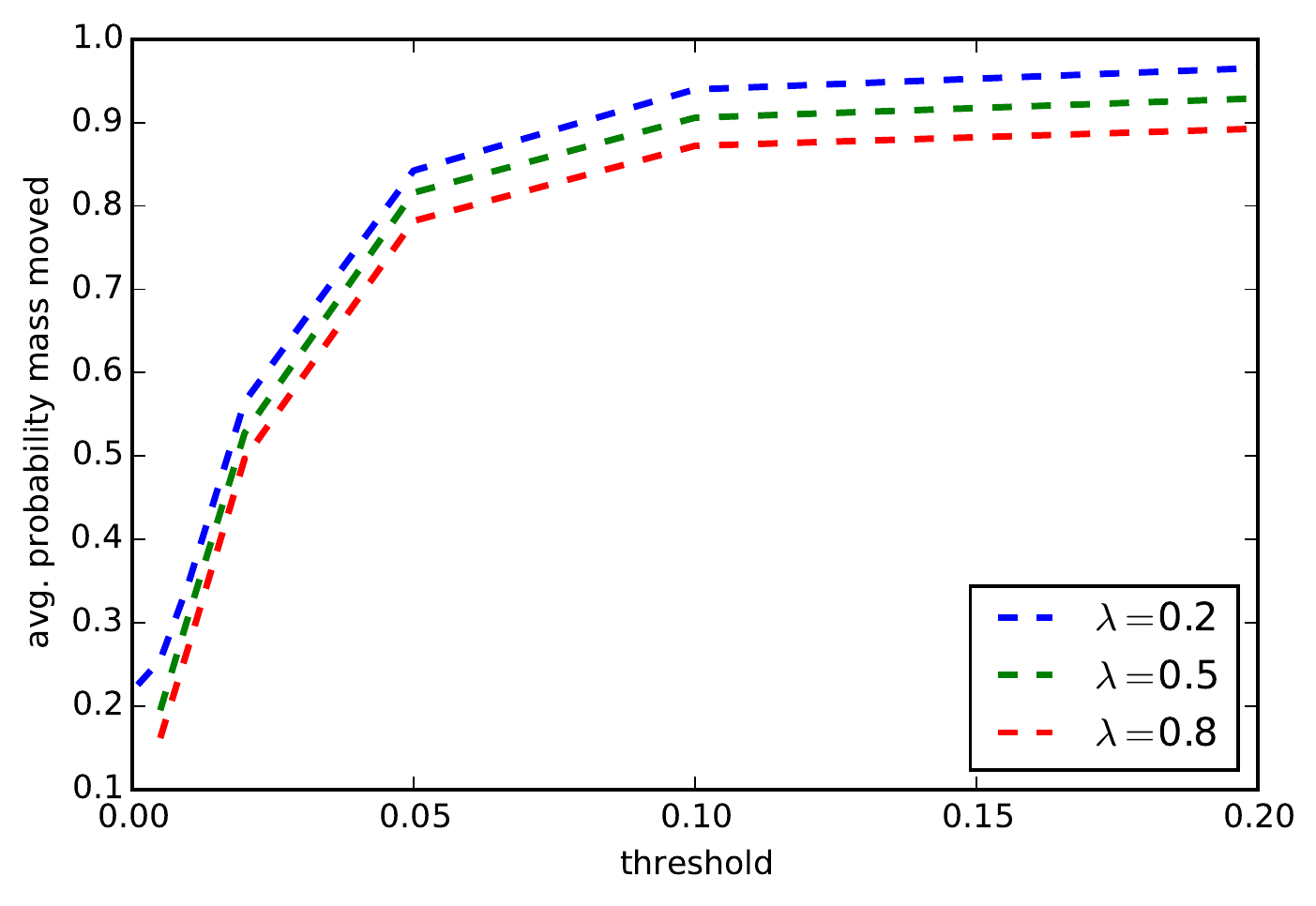} 
    \caption{Probability mass moved from removed words to the remaining words} 
    \label{fig7:a} 
    %\vspace{4ex}
  \end{subfigure}%% 
  \begin{subfigure}[b]{0.5\linewidth}
    \centering
    \includegraphics[width=0.75\linewidth]{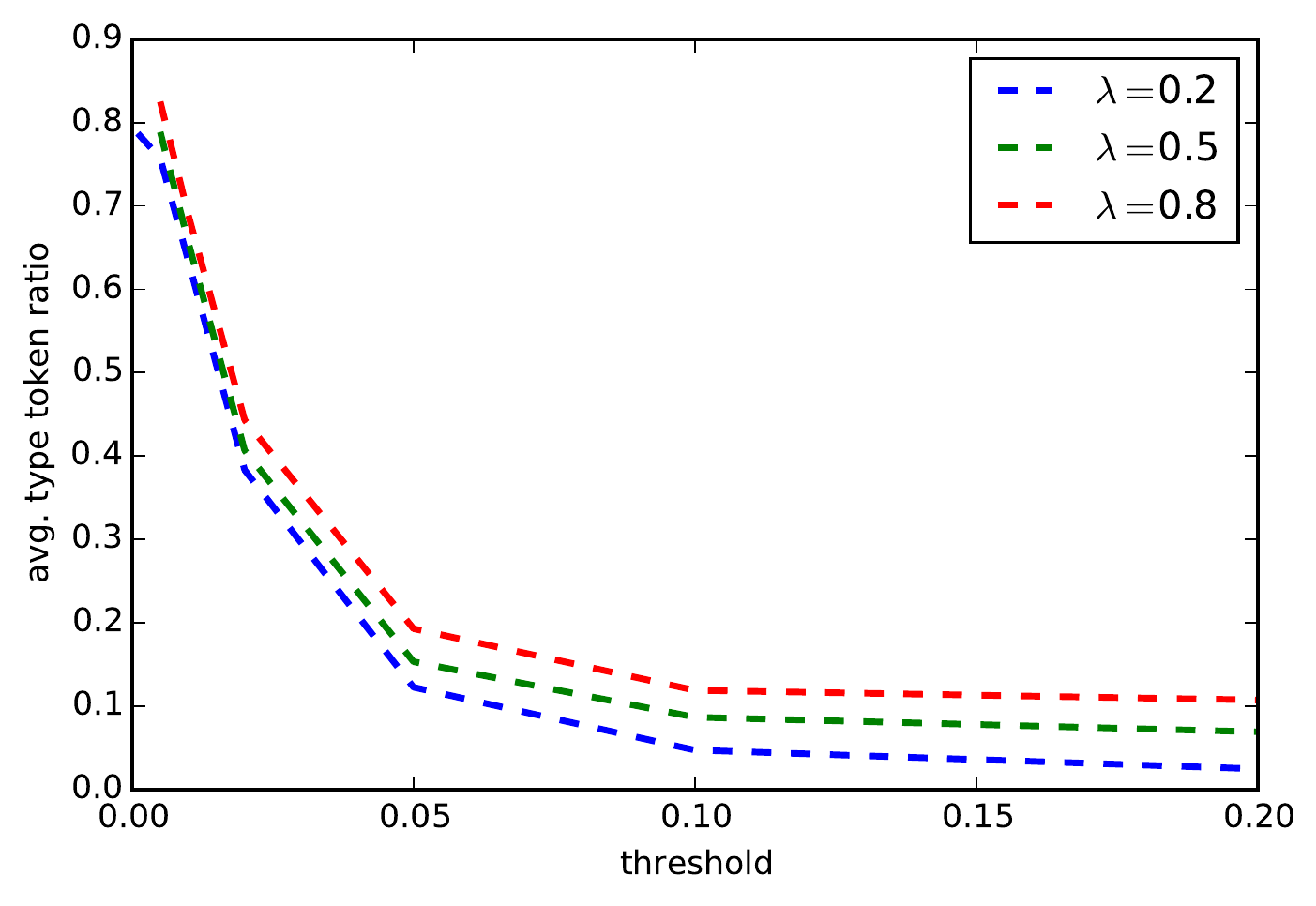} 
    \caption{Average type-token ratio of documents} 
    \label{fig7:b} 
    %\vspace{4ex}
  \end{subfigure} 
  \begin{subfigure}[b]{0.5\linewidth}
    \centering
    \includegraphics[width=0.75\linewidth]{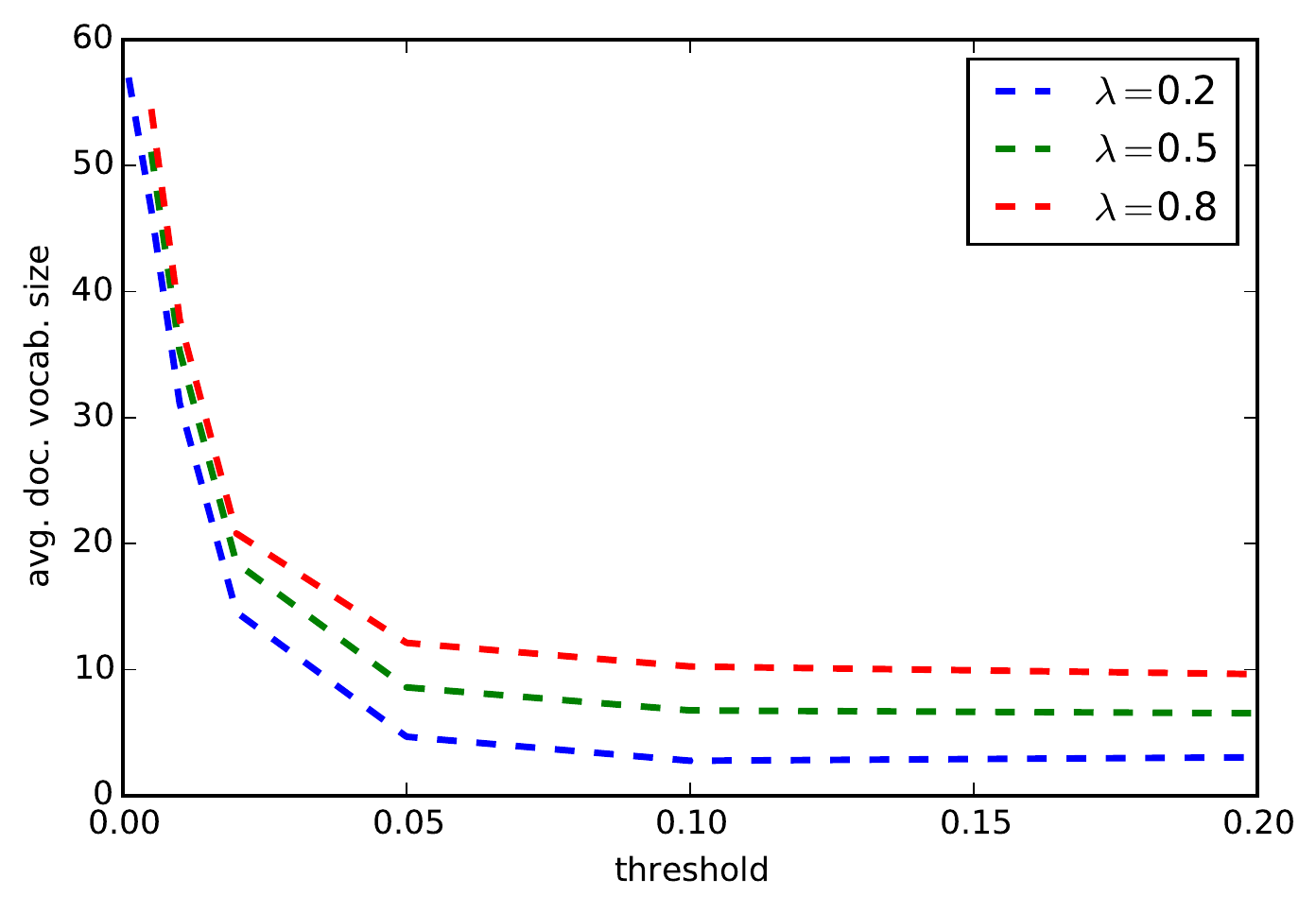} 
    \caption{Average document vocabulary size} 
    \label{fig7:c} 
  \end{subfigure}%%
  \begin{subfigure}[b]{0.5\linewidth}
    \centering
    \includegraphics[width=0.75\linewidth]{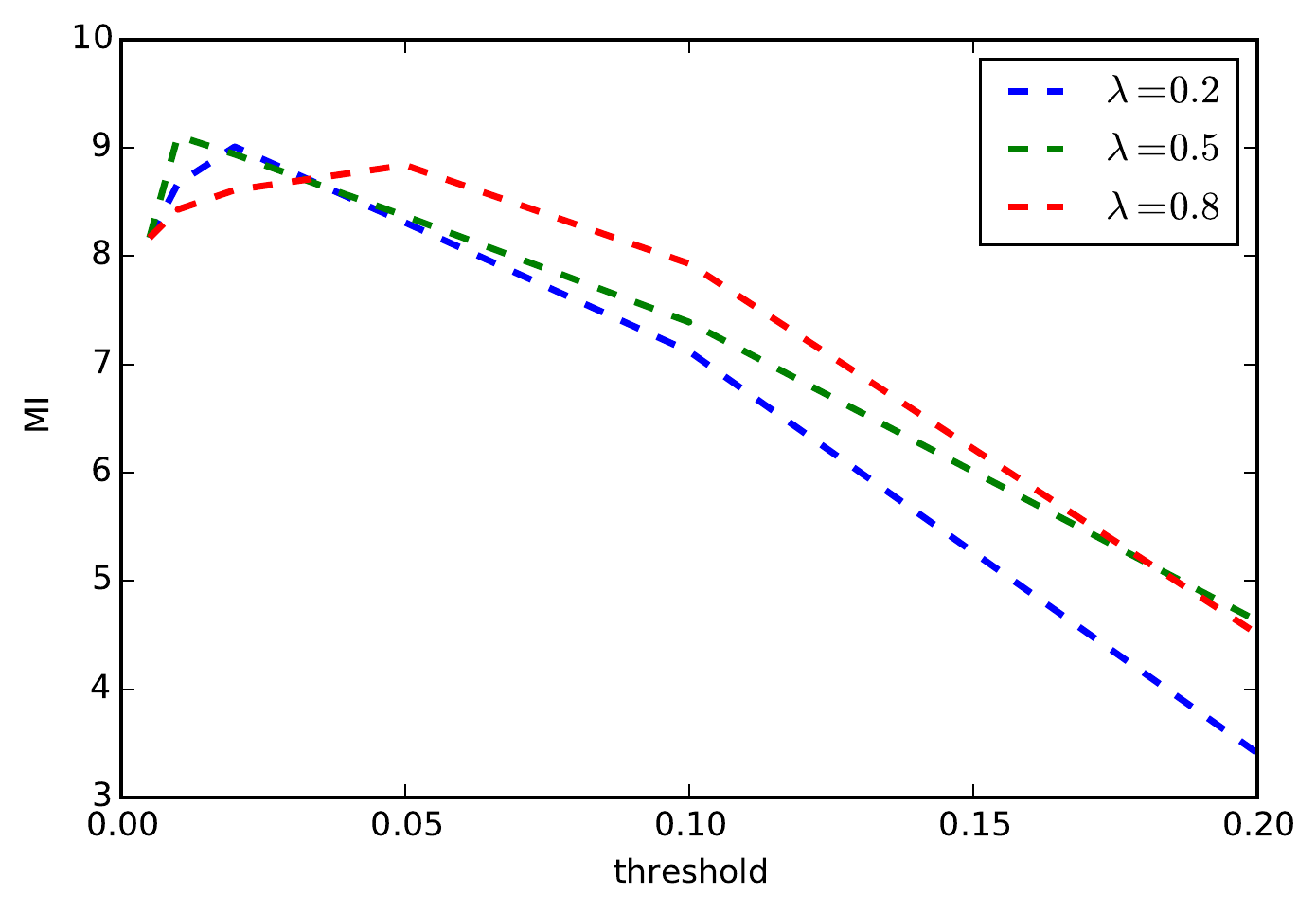} 
    \caption{Coherence of topic models estimated using Equation \ref{pmi-eq}} 
    \label{fig7:d} 
  \end{subfigure} 
  \caption{The effect of different values of the parameters of DR on the documents in terms of their probability mass moved, type-token ratio, and vocabulary size and its effect on the quality of trained topic models in terms of their coherence.}
  \label{fig-DR-param} 
\end{figure*}

Next, we analyze the effect of the amount of document re-estimation on the quality of topic models. We control the amount of re-estimation by the values of the parameters of DR: $\lambda$ and $threshold$. Fig.~\ref{fig-DR-param} shows the effect of different values of the parameters on documents and its impact on the quality of trained topic models. Two conclusions can be drawn. First, $\lambda$ does not have a great impact on the documents' statistics as even with very different values of $\lambda$ documents have similar statistics. The threshold has a bigger impact on the documents.  Second, although the statistics of documents are similar for different values of $\lambda$, the thresholds for which the best coherence is achieved for them, are very different. For $\lambda=0.5$ the best coherence is achieved for $threshold=0.01$, while for $\lambda=0.8$ the best coherence is achieved for $threshold=0.05$. This indicates that there is a correlation between these parameters. As expected, when $\lambda$ is high, which corresponds to less re-estimation, the threshold should be high to remove unnecessary words from documents.

\subsubsection{Topic re-estimation results}
\label{TRResults}
\if 0
\textcolor{green}{
\begin{itemize}
\item We want to evaluate how TR addresses the impurity issue with topic models. %and improves the quality of topic models. 
\item We want to answer the following RQ: Does TR increase the purity of topics? If so, how does using the more pure topics influence the performance in topical diversity task?
\end{itemize}
}

\textcolor{red}{
\begin{itemize}
\item A table including the mutual information between top 10 words within topics for LDA, LDA+TR, and LDA+DR+TR, and PTM. 
\item A table including topic before and after TR.\\
\item A histogram showing the probability mass of the words left after TR in the topics of the original LDA model.
\item A figure including the topic statistics (probability mass moved and average length of topics) using different parameters for TR and the coherence of the achieved topics in terms of mutual information.
\end{itemize}
}
\fi

To answer our third research question, we now focus on the TR level of HiTR. Since TR tries to remove the impurity from topics, we expect TR to increase the coherence of the topics by removing unnecessary words from topics. Table~\ref{table:4} shows the top five words for some example topics calculated from the PubMed dataset, before and after applying TR. 
These examples indicate that TR can successfully remove general words from topics.

\begin{table}[h]
	\centering 
		\caption{\label{table:4}
          Examples of topics before and after applying topic re-estimation on the PubMed dataset.} 
	\begin{tabular}{c l c l c}
    \toprule
    & \multicolumn{2}{c}{{Before TR}} & \multicolumn{2}{c}{{After TR}} \\
    \cmidrule(r){2-3}\cmidrule{4-5}
    Topic $t$ & $w$ & $p(w \conditional t)$ & $w$ & $p(w \conditional t)$ \\
    \midrule
    \multirow{5}{*}{1} & women & 0.07 & women & 0.06 \\
    & men & 0.02 & men & 0.05 \\
    & costs & 0.02 & health & 0.05 \\
    & per & 0.02 & costs & 0.03 \\
    & total & 0.02 & economic & 0.02 \\
    \midrule
    \multirow{5}{*}{2} & using & 0.01 & algorithm & 0.04 \\
    & method & 0.01 & method & 0.03 \\
    & used & 0.01 & data & 0.03 \\
    & algorithm & 0.01 & performance & 0.02 \\
    & data & 0.01 & system & 0.01 \\
    \midrule
    \multirow{5}{*}{3} & sequences & 0.02 & genome & 0.05 \\
    & genome & 0.02 & sequences & 0.04 \\
    & genes & 0.02 & genes & 0.03 \\
    & using & 0.01 & genomic & 0.03 \\
	 & two & 0.01 & gene & 0.02 \\
    \bottomrule
	\end{tabular}
\end{table}

We measure the purity of topics based on the coherence of words within $P(w\conditional t)$ distributions. Table~\ref{table:5} shows the coherence of topics according to different topic modeling approaches, in terms of average mutual information. 
More coherent topics are beneficial, because they are an indicator of more pure topics, which are essential to achieving a good performance in topical diversity task.  
TR increases the coherence of topics by removing the impure parts from topics. The coherence of PTM is higher than the coherence of TR. However, when we first apply DR, train LDA, and finally apply TR, the coherence of the extracted topics is significantly higher than the coherence of topics extracted by PTM.
From these findings we conclude that TR is effective in removing impurity from topics. Moreover, DR also contributes in making topics more pure.

\begin{table}[h]
	\centering
	\caption{\label{table:5}
	  The coherence of different topic models in terms of average mutual information between top 10 words in the topics calculated using Equation \ref{pmi-eq} on the PubMed dataset.} 
	\begin{tabularx}{0.8\columnwidth}{l C}
          \toprule	
          \textbf{Method} & \textbf{Coherence} \\
          \midrule 
          LDA &  8.17\\
          PTM &  9.89\\
          LDA+TR & 9.46\\
          LDA+DR+TR & 10.29$^\blacktriangle$\\
          \bottomrule
	\end{tabularx}
\end{table}	

\begin{figure}[h]
\centering
\includegraphics[width=\columnwidth]{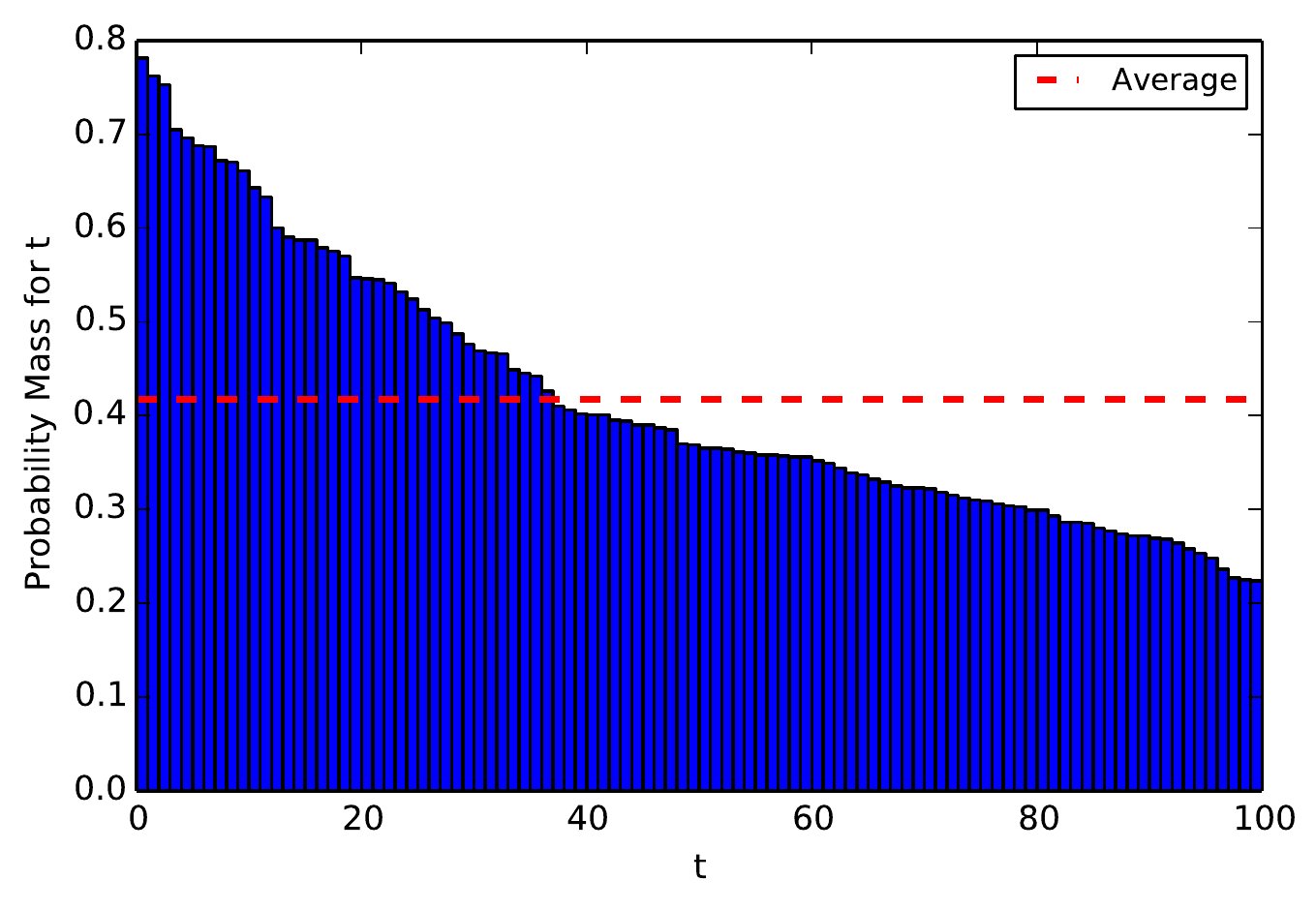}
\vspace{-2\baselineskip}
\caption{Probability mass of the words left after TR in the topics of the original LDA model. The y-axis shows $\sum_{\{w\mid P_{LDA+TR}(w\conditional t)>0\}} P_{LDA}(w\conditional t)$ for a topic $t$.}
\label{figure5}
\end{figure}

To see how much impurity is being removed from topics by using TR, we investigate the effect of TR on the distribution of words within topics and we measure the number of words and the re-allocated probability mass within topics before and after TR. 
Fig.~\ref{figure5} shows the probability mass of the words left after TP is applied to the topics of the original LDA model. 
The average number of words within extracted topics from the PubMed dataset is about 337 without TR, and about 181 after performing TR. On average, the words that are not removed by TR take 41\% of the probability mass in the LDA topic models (the dotted red line in Fig.~\ref{figure5}).
In the re-estimated topic model, they occupy the full 100\% of the probability mass.
 Thus, after applying TR, the topic models become more sparse, and the remaining topic-specific words receive higher probabilities. As shown in the figure, over all topics, after applying TR, the probability mass is re-allocated and some words are removed.
%We conclude that by re-allocating the probability mass within topics, TR removes impurity from topics.

\subsubsection{Topic assignment re-estimation results}
\label{TARResults}
\if 0
\textcolor{green}{
\begin{itemize}
\item We want to show the effect of TAR on the quality of topic distribution over documents. We measure it in terms of performance in topical diversity task. 
\item We want to evaluate if TAR is successful in addressing general topics issue which it is designed for. 
\item We want to analyze the effect of TAR on the sparsity of topic assignments to the documents.
\item We want to answer the following RQ: How does TAR affect the sparsity of document-topic assignments? And what is the effect of re-estimated document-topic assignments on the topical diversity task?
\end{itemize}
}

\textcolor{red}{
\begin{itemize}
\item A table including the sparsity of topic-assignments after applying TAR, compared to LDA and PTM. 
\item \todo{including median and standard deviation for sparsity.}
\item \todo{scatter plots/histograms of number of topics for PTM and HiTR}
\item A figure including the total probability of assigning topics to document for LDA and LDA+TAR. An analysis on for what kind of topics TAR decreases their total probability in the corpus.
\end{itemize}
}
\fi

\begin{figure}[h]
\centering
\includegraphics[width=\columnwidth]{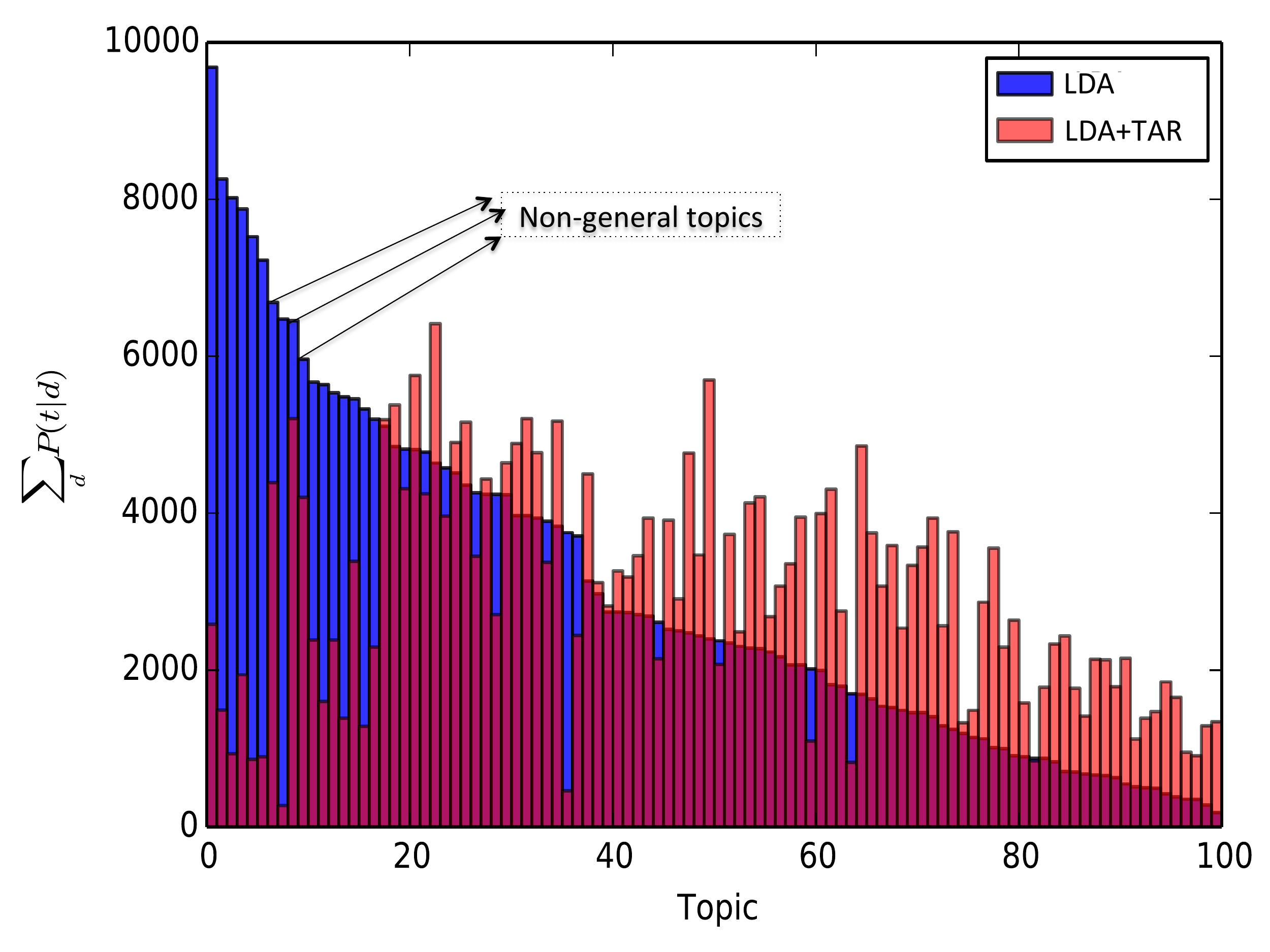}
\vspace*{-2\baselineskip}
\caption{The total probability of assigning topics to the documents in the PubMed dataset estimated using LDA and LDA+TAR.
(The two areas are both equal to the number of documents ($N\approx300K$)).} 
\label{figure4}
\end{figure}

To answer our fourth research question, we now turn to the TAR level of HiTR. We are interested in seeing how HiTR deals with the issue of general topics. General topics are topics that, for many documents, have a high probability of being assigned.
To gain insight in how LDA and HiTR perform in this respect, we sum the probability of assigning a topic to a document, over all documents: for each topic $t$, we calculate $\sum_{d \in C} P(t\conditional d)$, where $C$ is the collection of all documents. Fig.~\ref{figure4} shows the distribution of probability mass before and after applying TAR.  General topics naturally have high values as they are assigned to most of the documents with high probability. In Fig.~\ref{figure4} the topics are sorted based on the topic assignment probability of LDA.  As we can see from Fig.~\ref{figure4}, LDA assigns a vast portion of the probability mass to a relatively small number of topics. These topics are mostly general topics that are assigned to most of documents.  We expect, however, that many topics are represented in some documents, while relatively few topics will be relevant to all documents. When TAR is applied, the distribution is less skewed and the probability mass is more evenly distributed.

\noindent%
There are some topics that have high $\sum_d P(t\conditional d)$ value in LDA's topic assignments and high $\sum_d P(t\conditional d)$ value after applying TAR as well (they are marked as ``non-general topics'' in Fig.~\ref{figure4}).
Table~\ref{table:7} shows the top five words for these topics. Although these topics contain some general words such as ``used'', they are not general topics.
TAR is able to find these three non-general topics and their assignment probabilities to documents in the $P(t\conditional d)$ distributions is not changed as much as the actual general topics. 
%These results indicate that TAR removes general topics from documents and increases the probability of document-specific topics for each document.

\begin{table}[h]
	\centering 
	\caption{\label{table:7}
          Top five words for the topics marked as ``non-general topics'' in Fig.~\ref{figure4}.} 
          \resizebox{\columnwidth}{!}{%
	\begin{tabular}{c l}
    \toprule
    Topic & Top 5 words \\
    \hline
    1 & health, services, public, countries, data \\
    2 & surgery, surgical, postoperative, patient, performed \\
    3 & cells, cell, treatment, experiments, used \\
    \bottomrule
	\end{tabular}
	}
\end{table}

To further investigate whether TAR really removes general topics, in Table~\ref{table:6} we show the top five words for the first 10 topics in Fig.~\ref{figure4}, excluding the topics marked as ``non-general topics'' in the figure.
These seven topics have the highest decrease in  $\sum_d P(t\conditional d)$ values when we apply TAR.  As can be seen from Table~\ref{table:6}, the topics contain general words and are not informative.
In the figure, we can see that after applying TAR, the $\sum_d P(t\conditional d)$ values are decreased dramatically for these topics and that the mass is re-distributed across other topics, without creating new general topics that apply to nearly all documents.
We can conclude that TAR can correctly distinguish general from specific topics and re-assign probability mass accordingly. 

\begin{table}[h]
	\centering 
	\caption{\label{table:6}
          Top five words for the topics detected by TAR as general topics.} 
	\begin{tabular}{c l}
    \toprule
    Topic & Top 5 words \\
    \hline
    1 & use, information, also, new, one \\
    2 & ci, study, analysis, data, variables \\
    3 & time, study, days, period, baseline \\
    4 & group, control, significantly, compared, groups \\
    5 & study, group, subject, groups, significant \\
    6 & may, also, effects, however, would \\
    7 & data, values, number, average, used \\
    \bottomrule
    \end{tabular}
\end{table}

\subsection{Parameter analysis}
\begin{figure}[h]
\centering
\includegraphics[width=\columnwidth]{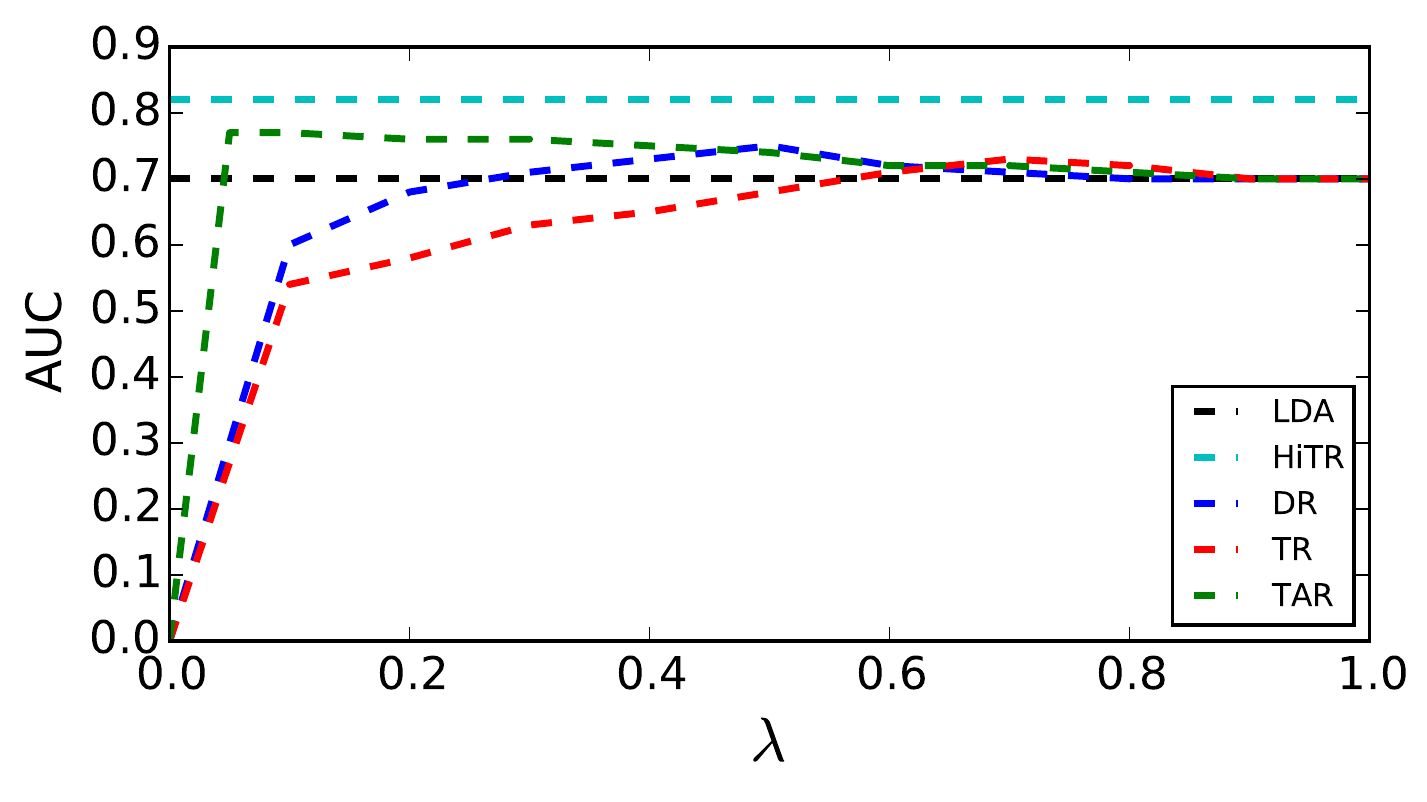}
\vspace*{-2\baselineskip}
\caption{The effect of the $\lambda$ parameter on the performance of topics models in topical diversity task on PubMed dataset.}
\label{figure2}
\end{figure}

In this section we analyze the effect of the $\lambda$ parameter on the performance of DR, TR, and TAR in the topical diversity task.
Fig.~\ref{figure2} displays the performance at different levels of re-estimation based on a range of values for $\lambda$. 
Recall that with $\lambda=1$, no re-estimation takes place, and all methods equal LDA.
The following interesting observations can be made from this figure.

First, DR reaches its best performance with moderate values of $\lambda$ ($0.4 \leq \lambda \leq 0.45$). This reflects that documents contain a moderate amount of general information and that DR is able to successfully deal with it. For $\lambda \geq 0.8$ the performance of DR and LDA is the same and for these values of $\lambda$ DR does not increase the quality of LDA.

Second, the best performance of TR is achieved with high values of $\lambda$ ($0.65 \leq \lambda \leq 0.75$). This indicates that topics usually only need a small amount of re-estimation. With this slight re-estimation, TR is able to improve the quality of LDA. However, for the values of $\lambda \geq 0.75$ the accuracy of TR degrades.

Third, TAR achieves its best performance with very low values of $\lambda$ ($0.02 \leq \lambda \leq 0.05$). These low values of $\lambda$ correspond to more re-estimation. From this result, we conclude that most of the noise is in the $P(t\conditional d)$ distributions, and that aggressive re-estimation allows TAR to remove most of this noise. 
The best values of $\lambda$ optimized for HiTR using the development set are close to the best values of $\lambda$ according to Fig.~\ref{figure2}.

\subsection{Impact of underlying topic model on the performance of HiTR}
In this section, we analyze the effect of using PTM as the underlying topic model for HiTR on the performance of HiTR. We apply HiTR on top of PTM and compare the results with the results of applying HiTR on top of LDA. Table \ref{table:hitrptm} shows the results of this experiment. The results show that: \begin{inparaenum}[(1)]
\item Applying HiTR on top of PTM does not improve PTM's performance significantly. We believe, the reason is that PTM already removes a lot of general information from topics/documents, but in some cases it also removes non-general information. LDA is in the other side of the spectrum, it keeps all information (general and non-general), and HiTR removes general information and keeps only the non-general information which leads to a higher performance. 
\item PTM benefits the most from the DR step. It shows that PTM is already effective in removing generality/impurity from topic-word and document-topic distributions, however it does not have a mechanism to remove generality/impurity from document-word distributions.
\item The performance of HiTR with LDA is significantly better than the performance of PTM and PTM with HiTR. As we mentioned, this shows that HiTR is more effective when the underlying topic model contains all information (general and non-general) and it can remove the non-general part. 
\item In terms of sparsity, HiTR makes PTM more sparse, however the difference is not significant. Thus, applying HiTR on an already sparse topic model does not have a big influence on its sparsity.
\end{inparaenum}

\begin{table}[h]
	\centering
	\caption{\label{table:hitrptm}
	  The performance and sparsity of HiTR using PTM as the underlying topic model in the topical diversity task.} 
	\begin{tabularx}{0.8\columnwidth}{l C C}
          \toprule	
          \textbf{Method} & \textbf{AUC} & \textbf{Sparsity} \\
          \midrule 
          PTM & 0.78 & 1.78\\
          \hline
          PTM+DR &  0.79 & 1.73 \\
          PTM+TR & 0.77 & 1.71 \\
          PTM+TAR & 0.78 & 1.65 \\
          PTM+HiTR & 0.79 & 1.63 \\
          \bottomrule
	\end{tabularx}
\end{table}

% !TEX root = ./main.tex

\section{Analysis}
\label{Analysis}
%In \S\ref{ExperimentalResults} we evaluated the performance of HiTR. We saw that it outperforms the state-of-the-art on the task for which it was designed: measuring topical diversity of text documents. 
In this section, we want to gain additional insights into HiTR and its effects on topic estimation. 
Purity of topic assignments to documents based on $P(t\conditional d)$ distributions has the highest effect on the quality of estimated diversity scores for documents. Therefore, it is important to measure how pure the estimated topic assignments are using HiTR.
In this section, we measure how much impurity is removed by HiTR from topic distributions.
 Then, we analyze the efficiency of HiTR. 
 
Based on the topics assigned by HiTR, LDA and PTM, we perform document clustering and document classification.
For clustering, following~\cite{Nguyen2015}, we consider each topic as a cluster. Each document $d$ is assigned to the topic that has the highest probability value in $P(t\conditional d)$.
For classification, we use all topics assigned to the document and consider them as features for a supervised classification algorithm. As the classification algorithm we use SVM. High accuracy achieved in document classification is then an indicator of high purity of topic distributions. 

We note that our focus in this section is not on achieving a top performance in document clustering and classification tasks: we only consider these tasks as a means to assess the purity of topic distributions using different topic models.

\subsection{Datasets}
We use three datasets: 20-NewsGroups,\footnote{Available at \small{\url{http://www.ai.mit.edu/people/~jrennie/20Newsgroups/}}} Reuters \cite{Lewis2004} and Ohsu\-med.\footnote{Available at \small{\url{http://disi.unitn.it/moschitti/corpora.htm}}}
The Reuters dataset contains 806,791 documents with category labels for 126 categories. For clustering and classification of documents, we use the 55 categories in the second level of the category hierarchy. 20-NewsGroups contains 20 categories and around 1,000 documents in each category, so in total there are about 20,000 documents. The Ohsumed dataset contains 50,216 documents grouped into 23 categories.

\subsection{Purity metrics}
For measuring the purity of clusters, two standard evaluation metrics are used: \emph{purity} and \emph{normalized mutual information} (NMI) \cite{Manning2008}.
%% In the document classification task, we use $P(t\conditional d)$ distributions as features so that we consider the probability of assigning a topic to a document as a feature for the document and train an SVM classifier on the training data, tune its parameters on the development set, and test its performance on the test data.
%% TK: We already said thta, and it's in the wrong subsection anyway 

\subsection{Settings}
We evaluate document clustering and classification using  10-fold cross validation and perform   the same document pre-processing as described in \S\ref{Preprocessing}.%  on documents before performing clustering and classification.

\subsection{Purity results}
\begin{table*}[t]
	\centering 
	\caption{\label{table:1}
Purity of topic models estimated in terms of purity achieved in document clustering. For significance tests, we consider {p-value} $<$ 0.05/7.} 	
          \begin{tabular}{l c c c c c c}
            %\toprule
    \toprule
    & \multicolumn{2}{c}{\textbf{Reuters (N=806,791, C=55)}} & \multicolumn{2}{c}{\textbf{20-Newsgroups (18,846, C=20)}} & \multicolumn{2}{c}{\textbf{Ohsumed (N=50,216, C=23)}} \\
   \cmidrule(r){2-3}\cmidrule(r){4-5}\cmidrule(r){6-7}
   \textbf{Method}& Purity  & NMI & Purity  & NMI & Purity  & NMI\\
   \midrule
   LDA & 0.55~~ & 0.40~~ & 0.52~~ & 0.36~~ & 0.50~~ & 0.30~~ \\
   %LDA+Entropy & 0.49 & 0.33 & 12.22 & 0.53 & 0.38 & 12.86 & 0.51 & 0.31 & 13.01 \\
   PTM & 0.61~~ & 0.43~~ & 0.57~~ & 0.38~~ & 0.55~~ & 0.33~~ \\
   \midrule
   LDA+DR & 0.57$^\blacktriangledown$ & 0.41$^\blacktriangledown$ & 0.56~~ & 0.39~~ & 0.53$^\blacktriangledown$ & 0.32$^\blacktriangledown$ \\
   LDA+TR & 0.57$^\blacktriangledown$ & 0.42$^\blacktriangledown$ & 0.56~~ & 0.38~~ & 0.53$^\blacktriangledown$ & 0.31$^\blacktriangledown$ \\
   LDA+TAR & 0.60~~ & 0.43~~ & 0.57~~ & 0.39~~ & 0.54~~ & 0.33~~ \\
   LDA+DR+TR & 0.58~~ & 0.42$^\blacktriangledown$ & 0.57~~ & 0.38~~ & 0.54~~ & 0.32~~ \\
   LDA+DR+TAR & 0.60~~ & 0.43~~ & 0.58~~ & 0.40~~ & 0.55~~ & 0.35$^\blacktriangle$ \\
   LDA+TR+TAR & 0.61~~ & 0.43~~ & 0.58~~ & 0.40$^\blacktriangle$ & 0.56$^\blacktriangle$ & 0.34$^\blacktriangle$ \\
   \midrule
   HiTR & \textbf{0.64}$^\blacktriangle$  &  \textbf{0.45}$^\blacktriangle$  & \textbf{0.60}$^\blacktriangle$  &  \textbf{0.42}$^\blacktriangle$  & \textbf{0.57}$^\blacktriangle$  &  \textbf{0.35} \\
   \bottomrule
	\end{tabular}
	%\vspace*{.3\baselineskip}
\end{table*}	

\begin{table*}[t]
	\centering 
	  \caption{\label{table:2}
Purity of topic models estimated in terms of accuracy achieved in document classification. For significance tests, we consider {p-value} $<$ 0.05/7. 
} 	
	  \begin{tabular}{l c c c c c c}
            \toprule	
            & \multicolumn{2}{c}{\textbf{Reuters (N=806,791, C=55)}} & \multicolumn{2}{c}{\textbf{20-Newsgroups (N=18,846, C=20)}} & \multicolumn{2}{c}{\textbf{Ohsumed (N=50,216, C=23)}} \\
               \cmidrule(r){2-3}\cmidrule(r){4-5}\cmidrule(r){6-7}
            \textbf{Method}& Accuracy  & Imp. over LDA & Accuracy & Imp. over LDA   & Accuracy & Imp. over LDA  \\
            \midrule
            LDA & 0.76~~ & -- & 0.81~~ & -- & 0.50~~ & --\\
            PTM & 0.82~~ & \phantom{0}8\% & 0.87~~ & \phantom{0}7\% & 0.56~~ & 12\% \\
            \midrule
            LDA+DR & 0.79$^\blacktriangledown$ & \phantom{0}4\% & 0.83$^\blacktriangledown$ & \phantom{0}2\% & 0.52$^\blacktriangledown$ & \phantom{0}4\% \\
            LDA+TR & 0.78$^\blacktriangledown$ & \phantom{0}3\% & 0.83$^\blacktriangledown$ & \phantom{0}2\% & 0.53$^\blacktriangledown$ & \phantom{0}1\% \\
            LDA+TAR & 0.82~~ & \phantom{0}8\% & 0.85$^\blacktriangledown$ & \phantom{0}5\% & 0.54~~ & \phantom{0}8\% \\
            LDA+DR+TR & 0.80$^\blacktriangledown$ & \phantom{0}5\% & 0.84$^\blacktriangledown$ & \phantom{0}4\% & 0.53$^\blacktriangledown$ & \phantom{0}6\% \\
            LDA+DR+TAR & 0.83~~ & \phantom{0}9\% & 0.86~~ & \phantom{0}6\% & 0.56~~ & 12\% \\
            LDA+TR+TAR & 0.82$^\blacktriangle$ & \phantom{0}8\% & 0.87~~ & \phantom{0}7\% & 0.58$^\blacktriangle$ & 16\% \\
            \midrule
            HiTR  &  \textbf{0.85}$^\blacktriangle$ & 12\% &  \textbf{0.89}$^\blacktriangle$ & 10\% &  \textbf{0.60}$^\blacktriangle$ & 20\% \\
	    \bottomrule
	  \end{tabular}
	%\vspace*{.4\baselineskip}
\end{table*}	

Table~\ref{table:1} shows the purity of HiTR in the document clustering task.
For all 3 datasets, on both measures, the purity of topics created by HiTR is significantly higher than with PTM.
%As we can see from the table, the topic distributions extracted using HiTR score higher than the ones extracted using LDA and PTM in terms of both purity and NMI. 
As expected, TAR is mostly responsible for the purity of $P(t\conditional  d)$: all runs which include TAR either improve or do not differ significantly from PTM. The different combinations show that also DR and TR yield additional purity, indicating that each of the three address different issues and contribute in a different way.

%This shows the ability of HiTR to make $P(t\conditional  d)$ more pure. 
%
%The two-level re-estimation topic models achieve higher purity values than their respective one-level counterparts except the combination of DR and TR. This indicates that re-estimation in each level contributes to the purity of $P(t\conditional  d)$. 
%The combination of TR and DR is not effective in increasing the purity values over its one-level counterparts on most of the datasets. This indicates that TR and DR address similar issues and their combination cannot increase the purity of $P(t\conditional  d)$. 
%However, when each of them are combined with TAR, the purity of the topic distributions increase. This shows that DR/TR and TAR are addressing different issues and both contribute in a different way.
%HiTR's ability in increasing the purity of $P(t\conditional  d)$ is due to taking the advantages of all three re-estimation methods in removing impurity.

Table~\ref{table:2} shows the performance of different topic models on the document classification task.
Again HiTR significantly outperforms PTM on all three datasets. We see the same trend as with clustering, but amplified: here all runs without TAR perform significantly worse than PTM. Note that on the smallest dataset, LDA and PTM performs already  well, and so are harder to improve.
 %
%HiTR is more accurate in estimating $P(t\conditional  d)$ and its accuracy is higher than that of other topic models.
Where in document clustering only the topics with the highest probability are considered, in document classification the classifiers use the entire $P(t\conditional d)$ distributions to classify documents. 
%\todo{I do not get the next sentence, "higher values of what?" what do you compare?  Isn't it the case that PTM is too aggresive, leaving only very few topics in P(t|d), so on clustering (where you just take the first topic) only the ranking and the winner is used, but here we use all of the distribution. So a bit more topics seems better. I don't know, try to make a convincing argument here. }
%
Performance of all methods in document classification is more closer to the perfect classifier than their performance in document clustering, as the maximum value of both accuracy and purity is 1. This indicates that the most probable topic does not necessarily contain all information about the content of a document.
In the cases that a document is about more than one topic, the classifier utilizes all $P(t\conditional d)$ information and performs better. 
Therefore, the higher accuracy of HiTR in this task is an indicator of its ability to assigning document-specific topics to documents. 

%\todo{I am not sure that what you see after this is a conclusion that you cabb drwa from this experiment.}
%Interestingly, unlike what we see while clustering documents, in the document classification case, PTM outperforms TAR.
%This indicates that TAR mostly tries to find the most probable topics of documents and it is not able to achieve pure topic-assignments when it is not combined with DR and TR. However, PTM focuses on finding accurate $P(t\conditional d)$ distributions for documents. 

\subsection{HiTR's efficiency}
Table~\ref{table:efficiency} shows the execution times of HiTR, LDA, and PTM. The reported execution time for HiTR is the time took to run HiTR once, given the corpus as input and topic assignments to documents as output. All models were run on machines with 6-core 3.0 GHz processors. The results show that, even on large datasets, HiTR does not add much complexity to LDA and the difference between the execution times of LDA and HiTR are reasonable. 
The execution times of PTM grow much faster than those of LDA and HiTR when the number of documents increase.

%\todo{Add sizes in words and nr of documents to this table and display the runtime in hours with 1 digit behind the comma. Also change the caption then.}
\begin{table}[t]
	\centering
	\caption{\label{table:efficiency}
	  The execution time of HiTR, LDA, and PTM in hours. $N$ and $\#w$ are the number of documents and tokens in the corpus, respectively.} 
	\begin{tabular}{c c c}
          \toprule	
          \textbf{Dataset} & \textbf{Method} & \textbf{Hours} \\
          \midrule 
          Reuters & LDA & 6.18 \\
          N=807K & PTM & 26.00~~ \\
          \#w=1,5M& HiTR & 9.17\\
          \midrule
          20-NewsGroups& LDA & 1.13 \\
          N=19K & PTM & 0.93 \\
          \#w=5,2M&HiTR & 1.45\\
          \midrule
          Ohsumed & LDA & 1.42 \\
          N=50K & PTM & 3.88 \\
          \#w=10M& HiTR & 2.45 \\
          \bottomrule
	\end{tabular}
\end{table}

\section{Conclusion}
%what have we done
We have proposed Hierarchical Topic model Re-estimation (HiTR), an approach for re-estimating topic models and applied them to measure topical diversity of text documents. 
%It addresses two main issues with topic models, topic generality and topic impurity, which negatively affect measuring topical diversity scores in three ways. 
%First, the existence of document-unspecific words within the distribution of words within documents yields general topics and impure topics. Second, the existence of topic-unspecific words within the distribution of words within topics yields impure topics. Third, the existence of document-unspecific topics within the distribution of topics within documents yields general topics. We have proposed three approaches for removing unnecessary or even harmful information from probability distributions, which we combine in our method for HiTR.

%what did we find and what does it mean
We have shown by experimental means that our approaches are able to remove general topics from topic models and increase the purity of topics. The results show that the estimated diversity scores for documents using HiTR are more accurate than those extracted using topic models created by LDA and PTM.
Our three main findings are as follows. First, general topics have the largest negative impact on the quality of topic models when they are used for measuring topical diversity. This indicates that purity of topic assignments is more important than purity of the distribution of words in topics and the distribution of words in documents in topical diversity task. The topic assignment re-estimation (TAR) that is designed to address this problem successfully detects general topics and removes them from documents. Second, re-estimation at each level helps to improve the quality of estimated diversity scores. We have shown that these "cleaned document topic models" yield better results when applied to measure topical diversity of documents. However, to achieve a highly accurate diversity scores, re-estimation at all three levels is needed to improve on the state-of-the-art PTM approach.
Third, we analyzed the effectiveness of HiTR in two other tasks: document clustering and document classification. We found that HiTR can achieve higher performances in these tasks compared to LDA and PTM. This finding suggest that although HiTR is originally designed for better estimation of topical diversity, it can be applied in a wider variety of tasks. 

%what are the limitations
Our proposed approach has some limitations. First, HiTR is most effective at removing general information from the probability distributions mentioned. However, to train a more accurate topic model which has a good performance in topical diversity task it is also important to remove very specific words from documents. Current approaches, including HiTR, are not able to address this problem adequately. 
%A possible direction is to design a re-estimation approach that not only removes general information from probability distributions but also removes very specific information from them. We hypothesize that this can be done in an effective manner by incorporating document frequency of words in the re-estimation process. 
Second, the experiments on the topical diversity task are conducted in an artificially created dataset. More robust datasets are needed for evaluating HiTR in this task. 

%what should we do next
There are several future directions. 
In principle, HiTR is a re-estimation method that can be applied to any topic model to enhance its quality. 
In this paper, we have applied HiTR to LDA and PTM. In our future work, we plan to examine the effect of HiTR on a wide range of topic models besides LDA and PTM such as PLSA.
%applying HiTR on probabilistic models such as images and pixels
In this research we adapted and used Rao's diversity measure for estimating diversity of documents. There are several other diversity measures proposed in biology such as Functional Divergence and Functional Attribute Diversity. 
%Another future direction would be employing other diversity measures for measuring topical diversity of text documents.

\section*{Acknowledgment}
This research was supported by the Netherlands Organization for Scientific Research (ExPoSe project, NWO CI \# 314.99.108; DiLiPaD project, NWO Digging into Data \linebreak \# 600.006.014), Nederlab (340-6148-t1-6), and by the European Community's Seventh Framework Program (FP7/2007-2013) under grant agreement ENVRI, number 283465.

\bibliographystyle{IEEEtranN}
\bibliography{IEEEabrv,ref}

% Generated by IEEEtranN.bst, version: 1.14 (2015/08/26)
\begin{thebibliography}{33}
\providecommand{\natexlab}[1]{#1}
\providecommand{\url}[1]{#1}
\csname url@samestyle\endcsname
\providecommand{\newblock}{\relax}
\providecommand{\bibinfo}[2]{#2}
\providecommand{\BIBentrySTDinterwordspacing}{\spaceskip=0pt\relax}
\providecommand{\BIBentryALTinterwordstretchfactor}{4}
\providecommand{\BIBentryALTinterwordspacing}{\spaceskip=\fontdimen2\font plus
\BIBentryALTinterwordstretchfactor\fontdimen3\font minus
  \fontdimen4\font\relax}
\providecommand{\BIBforeignlanguage}[2]{{%
\expandafter\ifx\csname l@#1\endcsname\relax
\typeout{** WARNING: IEEEtranN.bst: No hyphenation pattern has been}%
\typeout{** loaded for the language `#1'. Using the pattern for}%
\typeout{** the default language instead.}%
\else
\language=\csname l@#1\endcsname
\fi
#2}}
\providecommand{\BIBdecl}{\relax}
\BIBdecl

\bibitem[Bache et~al.(2013)Bache, Newman, and Smyth]{Bache2013}
K.~Bache, D.~Newman, and P.~Smyth, ``Text-based measures of document
  diversity,'' in \emph{Proceedings of the 19th ACM SIGKDD International
  Conference on Knowledge Discovery and Data Mining}, ser. KDD '13, 2013, pp.
  23--31.

\bibitem[Azarbonyad et~al.(2015)Azarbonyad, Saan, Dehghani, Marx, and
  Kamps]{Azarbonyad2015}
H.~Azarbonyad, F.~Saan, M.~Dehghani, M.~Marx, and J.~Kamps, ``Are topically
  diverse documents also interesting?'' in \emph{Proceedings of the 6th
  International Conference on Experimental IR Meets Multilinguality,
  Multimodality, and Interaction - Volume 9283}, ser. CLEF'15, 2015, pp.
  215--221.

\bibitem[Azarbonyad et~al.(2017)Azarbonyad, Dehghani, Kenter, Marx, Kamps, and
  de~Rijke]{Azarbonyad2017-ECIR}
H.~Azarbonyad, M.~Dehghani, T.~Kenter, M.~Marx, J.~Kamps, and M.~de~Rijke,
  ``Hierarchical re-estimation of topic models for measuring topical
  diversity,'' in \emph{Proceedings of the 39th European Conference on IR
  Research}, ser. ECIR '17, 2017, pp. 68--81.

\bibitem[Derzinski and Rohanimanesh(2014)]{Derezinski2015}
M.~Derzinski and K.~Rohanimanesh, ``An information theoretic approach to
  quantifying text interestingness,'' in \emph{Advances in Neural Information
  Processing Systems MLNLP workshop}, 2014.

\bibitem[Rao(1982)]{Rao1982}
C.~Rao, ``Diversity and dissimilarity coefficients: A unified approach,''
  \emph{Theoretical Population Biology}, vol.~21, no.~1, pp. 24--43, 1982.

\bibitem[Solow et~al.(1993)Solow, Polasky, and Broadus]{solow-measurement-1993}
A.~Solow, S.~Polasky, and J.~Broadus, ``On the measurement of biological
  diversity,'' \emph{Journal of Environmental Economics and Management},
  vol.~24, no.~1, pp. 60--68, 1993.

\bibitem[Soleimani and Miller(2015)]{Soleimani2015}
H.~Soleimani and D.~Miller, ``Parsimonious topic models with salient word
  discovery,'' \emph{IEEE Trans. on Knowl. and Data Eng.}, vol.~27, no.~3, pp.
  824--837, 2015.

\bibitem[Wallach et~al.(2009{\natexlab{a}})Wallach, Mimno, and
  McCallum]{Wallach2009}
H.~M. Wallach, D.~M. Mimno, and A.~McCallum, ``Rethinking lda: Why priors
  matter,'' in \emph{Advances in Neural Information Processing Systems}, ser.
  NIPS '09, 2009, pp. 1973--1981.

\bibitem[Lin et~al.(2014)Lin, Tian, Mei, and Cheng]{Lin2014}
T.~Lin, W.~Tian, Q.~Mei, and H.~Cheng, ``The dual-sparse topic model: Mining
  focused topics and focused terms in short text,'' in \emph{Proceedings of the
  23rd International Conference on World Wide Web}, ser. WWW '14, 2014, pp.
  539--550.

\bibitem[Hiemstra et~al.(2004)Hiemstra, Robertson, and Zaragoza]{Hiemstra2004}
D.~Hiemstra, S.~Robertson, and H.~Zaragoza, ``Parsimonious language models for
  information retrieval,'' in \emph{Proceedings of the 27th Annual
  International ACM SIGIR Conference on Research and Development in Information
  Retrieval}, ser. SIGIR '04, 2004, pp. 178--185.

\bibitem[Blei et~al.(2003)Blei, Ng, and Jordan]{Blei2003}
D.~M. Blei, A.~Y. Ng, and M.~I. Jordan, ``Latent dirichlet allocation,''
  \emph{Journal of Machine Learning Research}, vol.~3, pp. 993--1022, 2003.

\bibitem[Boyd-Gaber et~al.(2014)Boyd-Gaber, Mimno, and Newman]{Boyd2014}
J.~Boyd-Gaber, D.~Mimno, and D.~Newman, ``Care and feeding of topic models:
  Problems, diagnostics, and improvements,'' in \emph{Handbook of Mixed
  Membership Models and Their Applications}.\hskip 1em plus 0.5em minus
  0.4em\relax CRC Press, 2014.

\bibitem[Wang and Blei(2009)]{Wang2009}
C.~Wang and D.~M. Blei, ``Decoupling sparsity and smoothness in the discrete
  hierarchical dirichlet process,'' in \emph{Advances in Neural Information
  Processing Systems}, ser. NIPS '09, 2009, pp. 1982--1989.

\bibitem[Williamson et~al.(2010)Williamson, Wang, Heller, and
  Blei]{Williamson2010}
S.~Williamson, C.~Wang, K.~A. Heller, and D.~M. Blei, ``The {IBP} compound
  {D}irichlet process and its application to focused topic modeling,'' in
  \emph{Proceedings of the 27th International Conference on Machine Learning},
  ser. ICML '10, 2010, pp. 1151--1158.

\bibitem[Wallach et~al.(2009{\natexlab{b}})Wallach, Murray, Salakhutdinov, and
  Mimno]{Wallach2009b}
H.~M. Wallach, I.~Murray, R.~Salakhutdinov, and D.~Mimno, ``Evaluation methods
  for topic models,'' in \emph{Proceedings of the 26th Annual International
  Conference on Machine Learning}, ser. ICML '09, 2009, pp. 1105--1112.

\bibitem[Nguyen et~al.(2015)Nguyen, Billingsley, Du, and Johnson]{Nguyen2015}
D.~Q. Nguyen, R.~Billingsley, L.~Du, and M.~Johnson, ``Improving topic models
  with latent feature word representations,'' \emph{Transactions of the
  Association for Computational Linguistics}, vol.~3, pp. 299--313, 2015.

\bibitem[Lacoste-Julien et~al.(2009)Lacoste-Julien, Sha, and
  Jordan]{Lacoste2009}
S.~Lacoste-Julien, F.~Sha, and M.~I. Jordan, ``{DiscLDA}: Discriminative
  learning for dimensionality reduction and classification,'' in \emph{Advances
  in Neural Information Processing Systems}, ser. NIPS '09, 2009, pp. 897--904.

\bibitem[Xie and Xing(2013)]{Xie2013}
P.~Xie and E.~P. Xing, ``Integrating document clustering and topic modeling,''
  in \emph{Proceedings of the Twenty-Ninth Conference on Uncertainty in
  Artificial Intelligence}, 2013, pp. 694--703.

\bibitem[Mehrotra et~al.(2013)Mehrotra, Sanner, Buntine, and Xie]{Mehrotra2013}
R.~Mehrotra, S.~Sanner, W.~Buntine, and L.~Xie, ``Improving {LDA} topic models
  for microblogs via tweet pooling and automatic labeling,'' in
  \emph{Proceedings of the 36th International ACM SIGIR Conference on Research
  and Development in Information Retrieval}, ser. SIGIR '13, 2013, pp.
  889--892.

\bibitem[Lau et~al.(2014)Lau, Newman, and Baldwin]{Lau2014}
J.~H. Lau, D.~Newman, and T.~Baldwin, ``Machine reading tea leaves:
  Automatically evaluating topic coherence and topic model quality,'' in
  \emph{Proceedings of the 14th Conference of the European Chapter of the
  Association for Computational Linguistics}, ser. EACL '14, 2014, pp.
  530--539.

\bibitem[Chang et~al.(2009)Chang, Gerrish, Wang, Boyd-graber, and
  Blei]{Chang2009}
J.~Chang, S.~Gerrish, C.~Wang, J.~L. Boyd-graber, and D.~M. Blei, ``Reading tea
  leaves: How humans interpret topic models,'' in \emph{Advances in Neural
  Information Processing Systems}, ser. NIPS '09, 2009, pp. 288--296.

\bibitem[Newman et~al.(2011)Newman, Bonilla, and Buntine]{Newman2011}
D.~Newman, E.~V. Bonilla, and W.~Buntine, ``Improving topic coherence with
  regularized topic models,'' in \emph{Advances in Neural Information
  Processing Systems}, ser. NIPS '11, 2011, pp. 496--504.

\bibitem[Law et~al.(2004)Law, Figueiredo, and Jain]{Law2004}
M.~Law, M.~Figueiredo, and A.~Jain, ``Simultaneous feature selection and
  clustering using mixture models,'' \emph{IEEE Trans. on Pattern Analysis and
  Machine Intelligence}, vol.~26, no.~9, pp. 1154--1166, 2004.

\bibitem[Constantinopoulos et~al.(2006)Constantinopoulos, Titsias, and
  Likas]{Constantinopoulos2006}
C.~Constantinopoulos, M.~Titsias, and A.~Likas, ``Bayesian feature and model
  selection for gaussian mixture models,'' \emph{IEEE Trans. on Pattern
  Analysis and Machine Intelligence}, vol.~28, no.~6, pp. 1013--1018, 2006.

\bibitem[Dehghani et~al.(2016{\natexlab{a}})Dehghani, Azarbonyad, Kamps, and
  Marx]{Dehghani2016-ICTIR}
M.~Dehghani, H.~Azarbonyad, J.~Kamps, and M.~Marx, ``On horizontal and vertical
  separation in hierarchical text classification,'' in \emph{Proceedings of the
  2016 ACM International Conference on the Theory of Information Retrieval},
  ser. ICTIR '16, 2016, pp. 185--194.

\bibitem[Zhai and Lafferty(2001)]{Zhai2001}
C.~Zhai and J.~Lafferty, ``Model-based feedback in the language modeling
  approach to information retrieval,'' in \emph{Proceedings of the Tenth
  International Conference on Information and Knowledge Management}, ser. CIKM
  '01, 2001, pp. 403--410.

\bibitem[Zhang et~al.(2002)Zhang, Callan, and Minka]{Zhang2002}
Y.~Zhang, J.~Callan, and T.~Minka, ``Novelty and redundancy detection in
  adaptive filtering,'' in \emph{SIGIR}, 2002.

\bibitem[Dehghani et~al.(2016{\natexlab{b}})Dehghani, Azarbonyad, Kamps, and
  Marx]{Dehghani2016-CLEF}
M.~Dehghani, H.~Azarbonyad, J.~Kamps, and M.~Marx, ``Two-way parsimonious
  classification models for evolving hierarchies,'' in \emph{Proceedings of the
  7th International Conference of the CLEF Association}, ser. CLEF '16, 2016,
  pp. 69--82.

\bibitem[Blei(2012)]{Blei2012}
D.~M. Blei, ``Probabilistic topic models,'' \emph{Commun. ACM}, vol.~55, no.~4,
  pp. 77--84, 2012.

\bibitem[Pub()]{PubMed}
``National {C}enter for {B}iotechnology {I}nformation, {U}.{S}. {N}ational
  {L}ibrary of {M}edicine. {P}ubmed {C}entral {O}pen {A}ccess {I}nitiative.
  2010.''

\bibitem[R\"{o}der et~al.(2015)R\"{o}der, Both, and Hinneburg]{Roder2015}
M.~R\"{o}der, A.~Both, and A.~Hinneburg, ``Exploring the space of topic
  coherence measures,'' in \emph{Proceedings of the Eighth ACM International
  Conference on Web Search and Data Mining}, ser. WSDM '15, 2015, pp. 399--408.

\bibitem[Lewis et~al.(2004)Lewis, Yang, Rose, and Li]{Lewis2004}
D.~D. Lewis, Y.~Yang, T.~G. Rose, and F.~Li, ``Rcv1: A new benchmark collection
  for text categorization research,'' \emph{J. Mach. Learn. Res.}, vol.~5, no.
  Apr, pp. 361--397, 2004.

\bibitem[Manning et~al.(2008)Manning, Raghavan, and Sch\"{u}tze]{Manning2008}
C.~D. Manning, P.~Raghavan, and H.~Sch\"{u}tze, \emph{Introduction to
  Information Retrieval}.\hskip 1em plus 0.5em minus 0.4em\relax Cambridge
  University Press, 2008.

\end{thebibliography}

\end{document}